\documentclass[10pt]{article} 
\usepackage[accepted]{tmlr}

\usepackage{amsmath,amsfonts,bm}

\def\eqref#1{equation~\ref{#1}}

\def\1{\bm{1}}

\DeclareMathAlphabet{\mathsfit}{\encodingdefault}{\sfdefault}{m}{sl}
\SetMathAlphabet{\mathsfit}{bold}{\encodingdefault}{\sfdefault}{bx}{n}

\usepackage{CJKutf8}
\usepackage[utf8]{inputenc} 
\usepackage[T1]{fontenc}    
\usepackage[pagebackref,breaklinks,colorlinks]{hyperref}
\hypersetup{
    colorlinks=true,
    filecolor=magenta,      
    pdftitle={LLaVA-OneVision},
    pdfpagemode=FullScreen,
}
\usepackage{url}            
\usepackage{booktabs}       
\usepackage{amsfonts}       
\usepackage{nicefrac}       
\usepackage{microtype}      
\usepackage{xcolor}         

\usepackage{amsmath}
\usepackage{amsfonts}
\usepackage{amssymb}
\usepackage{wrapfig}
\usepackage{subcaption}
\usepackage{multirow}
 \usepackage{mathtools} 

\usepackage{verbatim}

\usepackage{anyfontsize}

\usepackage{microtype}
\usepackage{graphicx}
\usepackage{booktabs} 
\usepackage{multirow}
\usepackage{amsmath,amssymb}
\usepackage{booktabs}
\usepackage{caption,subcaption}

\usepackage{xcolor}
\definecolor{mygreen}{HTML}{3cb44b}
\definecolor{skyblue}{HTML}{beffff}
\definecolor{lightgreen}{HTML}{90ee90}

\usepackage{color, colortbl}

\definecolor{emerald}{rgb}{0.31, 0.78, 0.37}

\usepackage{tcolorbox}
\usepackage{enumitem}
\setitemize{itemsep=10pt,topsep=0pt,parsep=0pt,partopsep=0pt}
\usepackage{colortbl}

\usepackage{xcolor}
\definecolor{mygreen}{HTML}{3cb44b}
\colorlet{myyellow}{green!10!orange!90!}
\makeatletter

\usepackage{tikz}
\usetikzlibrary{arrows,shapes,snakes,automata,backgrounds,fit,petri}
\usepackage{adjustbox}

\newcommand{\RN}[1]{%
	\textup{\lowercase\expandafter{\it \romannumeral#1}}%
}
\usepackage{tabu}

\newcommand{\eg}[0]{\emph{e.g., }}

\newcommand{\beq}{\vspace{0mm}\begin{equation}}
\newcommand{\eeq}{\vspace{0mm}\end{equation}}
\newcommand{\beqs}{\vspace{0mm}\begin{eqnarray}}
\newcommand{\eeqs}{\vspace{0mm}\end{eqnarray}}
\newcommand{\barr}{\begin{array}}
\newcommand{\earr}{\end{array}}

\newcommand{\Vcal}{\mathcal{V}}

\usepackage{color, colortbl}
\definecolor{Gray}{gray}{0.93}

\usepackage{lipsum}

\usepackage{pifont}

\usepackage{makecell}

\usepackage{xcolor,amsmath}
\usepackage[linesnumbered,ruled,vlined]{algorithm2e}
\DontPrintSemicolon

\usepackage{xcolor}
\definecolor{mygreen}{HTML}{3cb44b}

\SetKwComment{Comment}{\color{green!50!black}\# }{}

\newcommand{\var}{\texttt}

\SetKwProg{Function}{def}{:}{}

\SetKwProg{For}{for}{:}{}
\SetKwProg{If}{if}{:}{}
\newcommand{\VarSty}[1]{\textnormal{\ttfamily\color{blue!90!black}#1}\unskip}

\usepackage{cleveref}
\usepackage{alltt}
\tcbuselibrary{most}
\tcbset{
  aibox/.style={
    width=\textwidth,
    top=10pt,
    colback=white,
    colframe=black,
    colbacktitle=black,
    enhanced,
    center,
    attach boxed title to top left={yshift=-0.1in,xshift=0.15in},
    boxed title style={boxrule=0pt,colframe=white,},
  }
}
\newtcolorbox{AIbox}[2][]{aibox,title=#2,#1}
\newlength\savewidth

\usepackage{xcolor}

\definecolor{ntured}{HTML}{D71440}
\definecolor{ntublue}{HTML}{181C62}
\definecolor{ntured}{HTML}{D71440}
\definecolor{ntublue}{HTML}{181C62}

\usepackage{hyperref}
\usepackage{url}

\newcommand{\tableCellHeight}{1}
\newcommand{\tabstyle}[1]{
  \setlength{\tabcolsep}{#1}
  \renewcommand{\arraystretch}{\tableCellHeight}
  \centering
  \small
}

\definecolor{citecolor}{HTML}{0071bc}

\definecolor{light-gray}{gray}{0.6}
\definecolor{front-color}{HTML}{F5FFFA}
\definecolor{tabhighlight}{HTML}{e5e5e5}
\definecolor{improvement}{RGB}{225,97,78}
\definecolor{mygreen}{HTML}{3cb44b}
\definecolor{Gray}{gray}{0.93}
\def\MethodName{\text{LLaVA-Video$_{~\texttt{SlowFast}}$}}

\def\DataName{\text{LLaVA-Video-178K}}

\def\ModelName{\text{LLaVA-Video}}

\title{LLaVA-Video: Video Instruction Tuning With Synthetic Data}

\author{\name Yuanhan Zhang \email yuanhan002@e.ntu.edu.sg \\
    \addr S-Lab, Nanyang Technological University 
    \AND
    \name Jinming Wu \email wjm\_18@bupt.edu.cn \\
    \addr BUPT 
    \AND
    \name Wei Li \email liwei.speech@bytedance.com \\
    \addr ByteDance
    \AND
    \name Bo Li \email libo0013@e.ntu.edu.sg \\
    \addr  S-Lab, Nanyang Technological University
    \AND
    \name Zejun Ma \email mazejun@bytedance.com \\
    \addr ByteDance
    \AND
    \name Ziwei Liu \email ziwei.liu@ntu.edu.sg \\
    \addr S-Lab, Nanyang Technological University 
    \AND
    \name Chunyuan Li \email lichunyuan24@gmail.com \\
    \addr ByteDance
}

\def\month{06}  
\def\year{2025} 
\def\openreview{\url{https://openreview.net/forum?id=EElFGvt39K}} 

\begin{document}
\maketitle


\begin{abstract}
The development of video large multimodal models (LMMs) has been hindered by the difficulty of curating large amounts of high-quality raw data from the web. To address this, we consider an alternative approach, creating a high-quality synthetic dataset specifically for video instruction-following, namely~\DataName{}. This dataset includes key tasks such as detailed captioning, open-ended question-answering (QA), and multiple-choice QA. By training on this proposed dataset, in combination with existing visual instruction tuning data, we introduce \ModelName{}, a new video LMM. Our experiments demonstrate that \ModelName{} achieves strong performance across various video benchmarks, highlighting the effectiveness of our dataset. We plan to release the dataset, its generation pipeline, and the model checkpoints.
\end{abstract}
\section{Introduction}

We are in an era where large-scale computing and data is crucial for multimodal learning~\citep{li2024multimodalsurvey}. A significant recent advancement was introduced by visual instruction tuning~\citep{liu2024visual}, which laid the foundation for building a general-purpose visual assistant. Notably, it proposed a data generation pipeline to create high-quality image-language instruction-following data. This pipeline has inspired subsequent researches~\citep{li2024llavaonevision,li2024llavanext-strong,li2024llavanext-ablations,lin2024vila} aimed at generating diverse image-language instruction data across various visual domains, accelerating the development of visual instruction tuning techniques.

Compared to the construction of image-language instruction-following data, obtaining high-quality video-language instruction-following data is challenging~\citep{damonlpsg2023videollama,li2024videochatchatcentricvideounderstanding}. First, sourcing high-quality videos is difficult. We need to find videos with significant temporal changes that provide more knowledge than what image-language data can offer. However, we have found that most videos in current video-language instruction-following datasets~\citep{chen2024sharegpt4video,zhang2024direct} are relatively static. Additionally, these videos are mostly trimmed based on scene changes, resulting in simplified plots. Such simplified video-language instruction-tuning data is inadequate for models to understand videos with complex narratives. Furthermore, current video-language instruction-following datasets often use a very sparse sampling rate for frame annotation. For instance, ShareGPT4Video~\citep{chen2024sharegpt4video} has an average sampling rate of 0.15, sometimes sampling only 2 frames from a 30-second video. 
This sparse sampling rate is effective in describing overall scenes but fails to capture detailed movements or changes in the video, resulting in hallucination  when detailed descriptions of the video are required.

To overcome these shortcomings, we introduce a comprehensive video instruction-tuning dataset named~\DataName, consisting of 178,510 videos ranging from 0 to 3 minutes. This dataset is enriched with detailed annotations, open-ended questions, and multiple-choice questions, developed through a combination of GPT-4o~\citep{openai2024gpt4o} and human efforts. It features four favorable properties: \textbf{$(i)$ Extensive Video Source:} We conduct a comprehensive survey on the video sources of exsiting video understanding datasets, and conclude 10 major video data sources, from which we start our video data collection by building a video pool.
Although there are over 40 video-language datasets, their video data are mainly sourced from 10 datasets~\citep{Zhou2017YouCookIID,xue2022hdvila,goyal2017something,caba2015activitynet,kay2017kinetics,sigurdsson2016charades,wang2023internvid,shang2019vidor,grauman2022ego4d,zhu2023languagebind}, covering a wide range of video domains, such as activities, cooking, TV shows, and egocentric views.  \textbf{$(ii)$ Dynamic Untrimmed Video Selection:} From these sources, we use several filtering logic to select the most dynamic videos from the video data pool. Notably, we select original, untrimmed videos to ensure plot completeness. \textbf{$(iii)$ Recurrent Detailed Caption Generation Pipeline with Dense Frame Sampling:} We propose a detailed video caption pipeline that operates recurrently, enabling us to generate detailed captions for videos of any length. This pipeline has three levels, each level of description represents a different time-range: from 10 seconds to the entire video length. It is recurrent as the historical description from any level serves as the context for generating new descriptions at any level. Additionally, we adopted a dense sampling strategy of one frame per second to ensure the sampled frames are rich enough to represent the videos. \textbf{$(iv)$ Diverse Tasks:} Based on the detailed video descriptions, we can generate question-answer pairs. To ensure our questions cover a wide range of scenarios, by referring to the video question-answering dataset, we define 16 question types. We prompt GPT-4o to generate question-answer pairs by referring to these question types, covering open-ended and multi-choice questions.

Based upon the~\DataName~dataset, we developed~\ModelName. Contrary to previous studies suggesting that training with single frames is sufficient for video-language understanding~\citep{lei2022revealing}, our findings reveal a significant impact of frame count on~\ModelName's performance, attributable to the detailed features of~\DataName. Observing this, we explored maximizing frame sampling within the constraints of limited GPU memory. We introduce~\MethodName, a video representation technique that optimally distributes visual tokens across different frames. This approach allows for incorporating up to three times more frames than traditional methods, which allocate an equal number of visual tokens to each frame.

Our contributions are as follows:
\begin{itemize}[leftmargin=7.5mm]
\setlength{\itemsep}{2pt}
\item {\it Video-language Instruction-Following Data}: We present a high-quality dataset \textit{\DataName} tailored for video instruction-following. It consists of 178K video with 1.3M instruction samples, including detailed captions, free-form and multiple-choice question answering.

\item {\it Video Large Multimodal Models}: We develop \textit{\ModelName}, a series of advanced large video-language models that expand the capabilities of open models in understanding video content.

\item {\it Open-Source}: In an effort to support the development of general-purpose visual assistants, we release our multimodal instruction data, codebase, model checkpoints, and a visual chat demo to the public.
\end{itemize}

\section{Related Work}


\begin{figure*}[t!]
    \centering
    \includegraphics[width=0.95\textwidth]{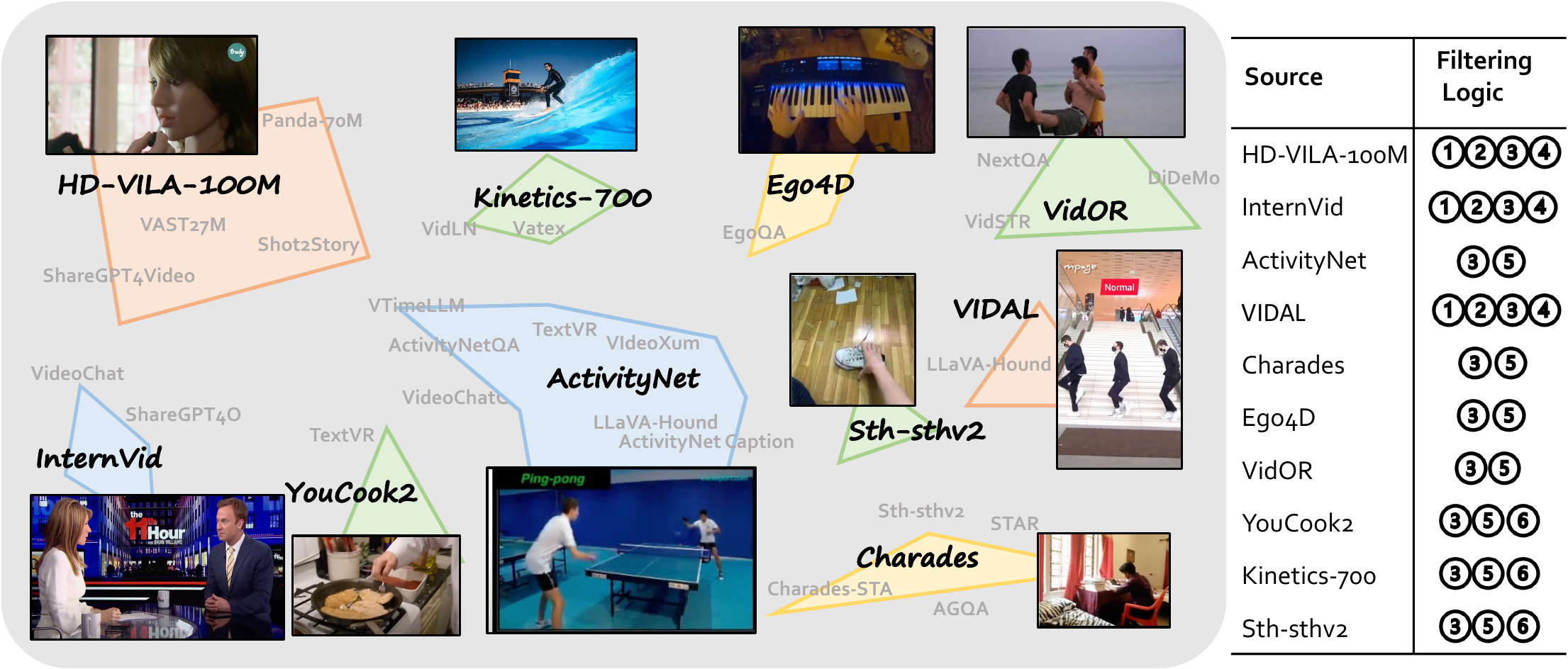}



    \caption{\textbf{Video sources in the proposed \textit{\DataName{}}}. (Left) The relationship between 10 video sources we have utilized and other existing video-language datasets. (Right) Filtering logic for video sources. The detail of filtering logic: \ding{172} Sorted by Views, \ding{173} Number of scenes greater than 2, \ding{174} Video duration between 5 seconds and 180 seconds, \ding{175} Ratio of scenes to video duration less than or equal to 0.5, \ding{176} Resolution greater than 480p, \ding{177} 50 samples for each category.}
\label{fig:video_data_source}
\end{figure*}

In this work, our goal is to create a high-quality video-language dataset that goes beyond simple video captions. We aim to improve the ability to follow instructions, which includes detailed video descriptions, open-ended video question-answering, and multiple-choice video question-answering data. We discuss related datasets in Table~\ref{tab:dataset_comparison}. Previous video-language datasets~\citep{miech19howto100m} include manually annotated data for various tasks, such as video captions~\citep{chen2011msvd,xu2016msrvtt,rohrbach2015LSMDC,anne2017didemo,caba2015activitynet,Zhou2017YouCookIID}, and video question-answering~\citep{yu2019activityqa,zadeh2019social,xiao2021next}. However, manual annotation is expensive and limits the size of such datasets. To address the shortage of data, studies like~\citep{miech19howto100m,lee2021acav100m,zellers2021merlot,xue2022hdvila} suggest automatically annotating data using subtitles created by ASR. While this method greatly expands the dataset size to 100 million samples, the subtitles often fail to accurately describe the main video content. Additionally, other studies~\citep{xu2017msvdqa,grunde2021agqa,wu2024star} use language models~\citep{xu2017msvdqa} or question templates~\citep{grunde2021agqa,wu2024star} to generate question-answer pairs. Although this approach can generate a large number of questions and answers, it often produces poor-quality questions that do not reflect real-world user inquiries. More recent research~\citep{chen2024panda70m} has prompted video-language models such as BLIP-2~\citep{li2023blip2bootstrappinglanguageimagepretraining}, VideoChat~\citep{li2024videochatchatcentricvideounderstanding}, Video-LLaMA~\citep{damonlpsg2023videollama}, and MiniGPT-4~\citep{zhu2023minigpt} to generate video captions. However, these models are limited in their ability to provide detailed descriptions. 

The most related works to ours are the recent AI-generated synthetic video instruction tuning data, \citet{islam2024video} introduced Video ReCap, which recursively annotates video captions. Unlike Video ReCap, each clip-wise (level-1) description in our pipeline is generated with historical context. This ensures that connections from previous events in the video timeline are linked to the current event. LLaVA-Hound~\citep{zhang2024direct} and ShareGPT4Video~\citep{chen2024sharegpt4video}, where they have used GPT-4~\citep{openai2023gpt4v} to generate video captions and open-ended video question-answering. Although the quality of the captions and question-answer pairs has significantly improved, the video sources they use are too static to produce high-quality data for instruction-following scenarios. They also only use very sparse frames for prompting GPT-4V, which results in annotations that fail to capture nuanced actions and continuous plots in the videos. Additionally, Shot2Story~\citep{han2023shot2story20k} and Vript~\citep{han2023shot2story20k} also employ GPT-4V~\citep{openai2023gpt4v} for video captioning. Their outputs, however, include audio details, which are outside the scope of this study.

\section{Video Instruction-Following Data Synthesis}
\label{sec:dataset}

A high-quality dataset for video instruction-tuning is crucial for developing effective video-language models. We identify a key factor in building such datasets: ensuring richness and diversity in both video content and its language annotations. We perform comprehensive survey on the existing video benchmarks, covering across various public video captioning and question-answering datasets, then identify ten unique video sources that contribute to over 40 video-language benchmarks. From each source, we select videos that exhibit significant temporal dynamics. To maintain diversity in the annotations, we establish a pipeline capable of generating detailed captions for videos of any length. Additionally, we define 16 types of questions that guide GPT-4o in creating question-answer pairs to assess the perceptual and reasoning skills of the video-language models.

\subsection{Video source}
\label{sec:dataset;subsec:datasource}
One important starting point in building a high-quality video instruction-following dataset is to find a sufficiently diverse pool of video data. From this pool, we can select the qualified videos. In our study of public video-language datasets---including video captioning, video question answering, video summarization, and moment-wise captioning---we noticed that although different datasets focus on various video understanding tasks (\eg, AGQA~\citep{grunde2021agqa} for spatial-temporal relations and STAR~\citep{wu2024star} for situational reasoning), most are sourced from ten main video sources. For instance, both AGQA and STAR use data from Charades~\citep{sigurdsson2016charades}. Specifically, these ten sources are HD-VILA-100M~\citep{xue2022hdvila}, InternVid-10M~\citep{wang2023internvid}, VidOR~\citep{shang2019vidor}, VIDAL (YouTube Shorts)\citep{zhu2023languagebind}, YouCook2\citep{Zhou2017YouCookIID}, Charades~\citep{sigurdsson2016charades}, ActivityNet~\citep{caba2015activitynet}, Kinetics-700~\citep{kay2017kinetics}, Something-Something v2~\citep{goyal2017something}, and Ego4d~\citep{grauman2022ego4d}. These sources offer a wide range of video data from different websites, viewpoints, and domains. The relationship between these ten selected video datasets and others is shown in Fig.~\ref{fig:video_data_source}. The videos from this ten datsets build the video pool for the further video selection.  Notably, we use untrimmed videos from each source except for YouCook2 and Kinetics-700. We believe that cutting videos into clips can break the plot continuity, which is essential for understanding the videos.

Based on the video pool, we aim to select dynamic videos. In Figure~\ref{fig:video_data_source}, we outline our criteria for selecting high-quality data. Our main method for identifying dynamic content involves using PySceneDetect, which calculates the number of scenes in a video
We found that the number of scenes is a good indicator of video dynamism. Additionally, we have designed a specific approach~\ding{175} to exclude videos that mainly contain ``slides."

\begin{figure*}[t!]
\centering
\includegraphics[width=0.95\textwidth]{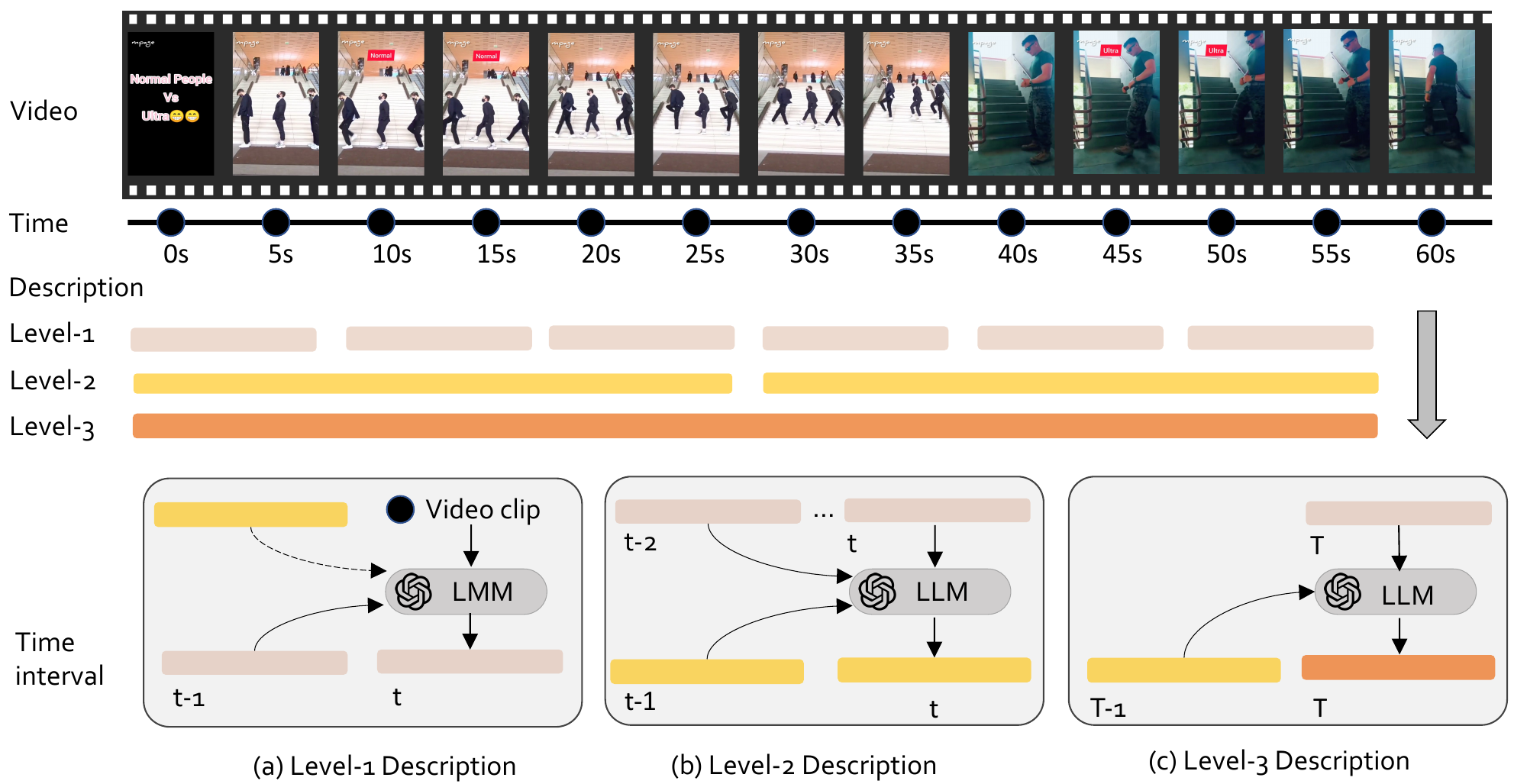}
\caption{\textbf{The video detail description creation pipeline}. A three-level creation pipeline is considered, with each level developed via a recurrent approach. 
Note that $t$ is the index of time internal at its own level, and $T$ is the last time internal index. 
(a) To generate the caption for time internal $t$ at level-1, we condition on the current frames in this internal, the caption for time internal $t-1$, and the most recent description summary at level-2 if applicable. 
(b) To generate caption for time internal $t$ at level-2, we condition on the previous caption at level-2, and captions from three most recent time internals at level-1.
(c) To generate the overall caption at the last time internal $T$ at level-3, we condition on the the most recent caption at level-2 and the current caption from level-1.
%
}
\label{fig:data_generation}
\end{figure*}

\subsection{Video Detail Description}
\label{sec:dataset;subsec:video_detail_description}
\paragraph{Automated Generation}
For selected videos, we use GPT-4o~\citep{openai2024gpt4o} to systematically describe their content. We start by sampling video frames at one frame per second (fps). However, due to the input size constraints of GPT-4o, we cannot use all sampled frames. Instead, we describe the videos sequentially, as shown in Fig~\ref{fig:data_generation}. We create descriptions at three distinct levels, detailed below.

\begin{itemize}[leftmargin=7.5mm]
\setlength{\itemsep}{2pt}
\item {\it Level-1 Description}: Every 10 seconds, we provide a level-1 description that outlines the events in that segment. This description considers: frames from the current clip and historical context, which includes all recent level-1 descriptions not yet summarized into a level-2 description and the latest level-2 description.

\item {\it Level-2 Description}: Every 30 seconds, we creat a level-2 summary of the entire video plot up to that point. This is based on the last three level-1 descriptions, covering the most recent 30 seconds; and the latest level-2 description.

\item {\it Level-3 Description}: At the video's end, we generate a level-3 description to encapsulate the entire video. The inputs for this description are the recent level-1 descriptions not yet summarized, covering the last moments of the plot after the recent summary; and the latest level-2 description.
\end{itemize}

\begin{figure*}[t!]
    \centering
    \includegraphics[width=0.95\textwidth]{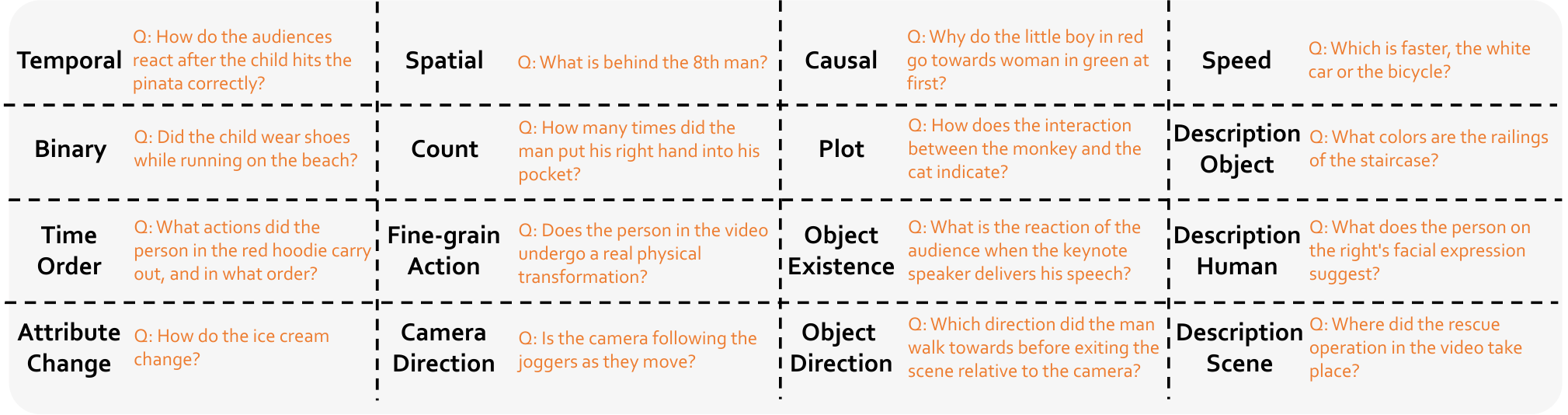}
    \caption{Question types for video question answering in data creation. For each type, we provide its name and an example question.}
 \label{fig:question_type}
\end{figure*}

\subsection{Video Question Answering}
\paragraph{Question Type definition}
In addition to detailed video descriptions, our dataset includes a variety of question-answer pairs designed for complex interactions. This setup improves the video understanding model's ability to handle real-life queries. We refer to public video question-answering benchmarks~\citep{xiao2021next,yu2019activityqa,Khattak2024cvrres,liu2024tempcompass} to organize these questions into 16 specific categories, as shown in Fig.~\ref{fig:question_type}.

\paragraph{Automated Generation}
Given a detailed video description, we use GPT-4o to generate at most one question-answer pair for each type of question. The prompts include: (1) The task definition for the current question type. (2) In-context examples for this type, which include three video descriptions and their three question-answer pairs of this specific type. (3) The detailed video description for the current video.
We instruct GPT-4o to return \textit{None} if it cannot generate question-answer pairs for a specific question type.

\paragraph{Filtering.}
To filter out the generated question-answer pairs, we apply the following strategy: (1) remove duplicates using the sentence-transformer~\citep{reimers-2020-multilingual-sentence-bert}, (2) discard answers that begin with phrases like ``does not specify,'' ``does not mention,'' ``does not specifically,'' ``does not depict,'' or ``does not show.''

\begin{figure*}[t]
\centering
\includegraphics[width=0.95\textwidth]{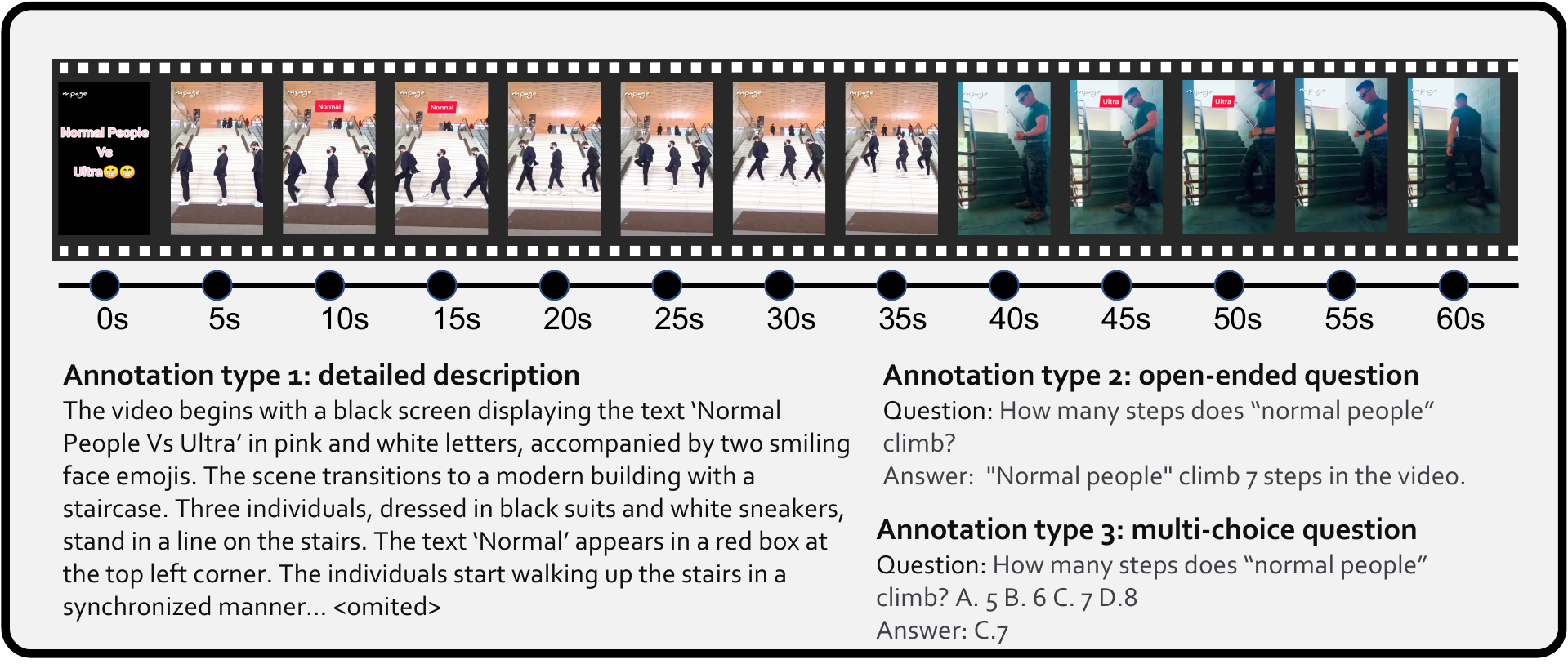}
\caption{One example to illustrate the video instruction-following data.}
\label{fig:annotation_example}
\end{figure*}

\subsection{Dataset Statistics}
\paragraph{Overview.} 
We carefully select from our collected data sources to form a balanced and comprehensive collection, resulting in a total of 178K videos and 1.3M instruction-following samples. This includes 178K captions, 960K open-ended QAs, and 196K multiple-choice QAs.

\begin{figure*}[t]
\begin{minipage}[b]{0.95\textwidth}
\centering
    \includegraphics[width=0.85\textwidth]{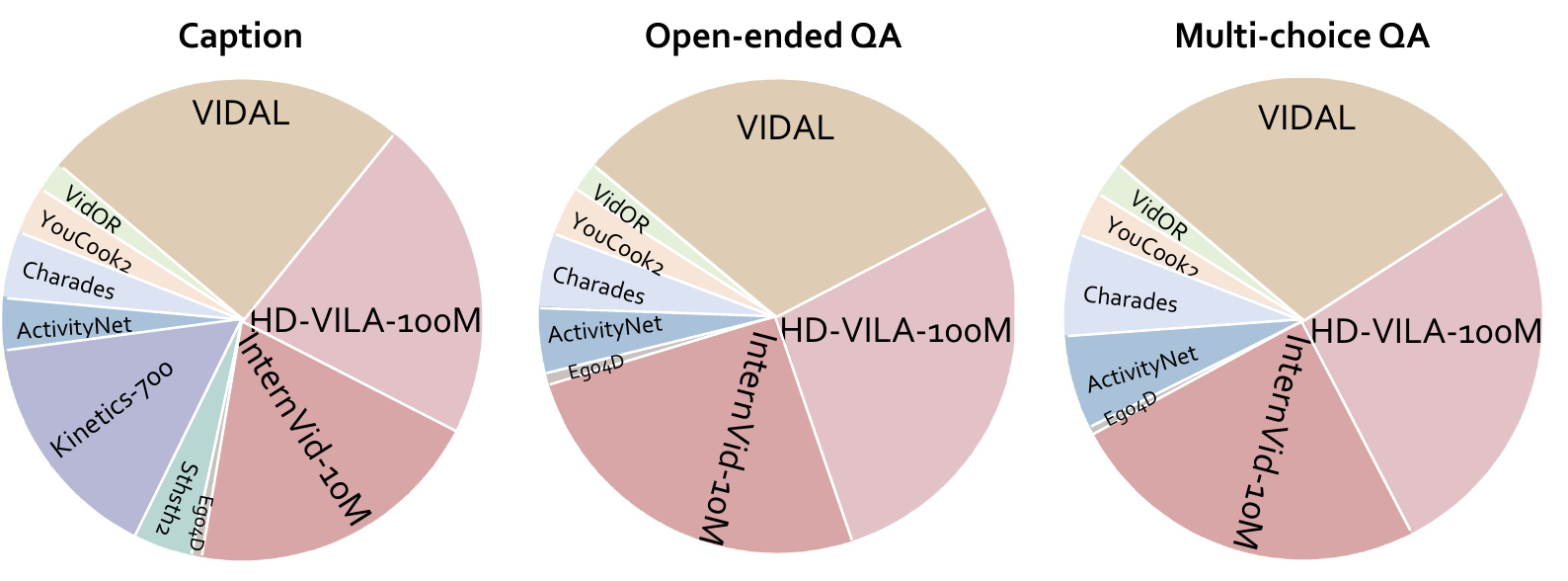}
\end{minipage}

\begin{minipage}[b]{0.95\textwidth}
\centering
\tabstyle{2pt}
\fontsize{8pt}{10pt}\selectfont
\begin{tabular}{l|ccc||l|ccc}
    \midrule
    \textbf{Dataset} &
    \textbf{\#Caption} &
    \textbf{\#Open-Ended} &
    \textbf{\#Multi-Choice} &
    \textbf{Dataset} &
    \textbf{\#Caption} &
    \textbf{\#Open-Ended} &
    \textbf{\#Multi-Choice}\\
    \midrule
    \makecell[l]{VidOR} &
    \tikz[baseline=0.05em] \fill [color={rgb,255: red,229; green,240; blue,219}] (0,0) rectangle (0.75em,0.75em); 4,018 &
    \tikz[baseline=0.05em] \fill [color={rgb,255: red,247; green,230; blue,216}] (0,0) rectangle (0.75em,0.75em); 19,875 &
    \tikz[baseline=0.05em] \fill [color={rgb,255: red,220; green,227; blue,242}] (0,0) rectangle (0.75em,0.75em); 4,773 & 
    \makecell[l]{Sthsth2} &
    \tikz[baseline=0.05em] \fill [color={rgb,255: red,185; green,215; blue,210}] (0,0) rectangle (0.75em,0.75em); 8,700 & 
    \tikz[baseline=0.05em] \fill [color={rgb,255: red,200; green,194; blue,190}] (0,0) rectangle (0.75em,0.75em); 0 &
    \tikz[baseline=0.05em] \fill [color={rgb,255: red,216; green,166; blue,166}] (0,0) rectangle (0.75em,0.75em); 0 \\
    \makecell[l]{YouCook2} &
    \tikz[baseline=0.05em] \fill [color={rgb,255: red,169; green,193; blue,217}] (0,0) rectangle (0.75em,0.75em); 7,411 &
    \tikz[baseline=0.05em] \fill [color={rgb,255: red,182; green,184; blue,214}] (0,0) rectangle (0.75em,0.75em); 32,143 &
    \tikz[baseline=0.05em] \fill [color={rgb,255: red,185; green,215; blue,210}] (0,0) rectangle (0.75em,0.75em); 5,776 &
    \makecell[l]{Ego4D} &
    \tikz[baseline=0.05em] \fill [color={rgb,255: red,226; green,194; blue,198}] (0,0) rectangle (0.75em,0.75em); 1,065 &
    \tikz[baseline=0.05em] \fill [color={rgb,255: red,222; green,205; blue,180}] (0,0) rectangle (0.75em,0.75em); 5,912 &
    \tikz[baseline=0.05em] \fill [color={rgb,255: red,229; green,240; blue,219}] (0,0) rectangle (0.75em,0.75em); 520 \\
    \makecell[l]{Charades} &
    \tikz[baseline=0.05em] \fill [color={rgb,255: red,200; green,194; blue,190}] (0,0) rectangle (0.75em,0.75em); 9,803 &
    \tikz[baseline=0.05em] \fill [color={rgb,255: red,216; green,166; blue,166}] (0,0) rectangle (0.75em,0.75em); 48,187 &
    \tikz[baseline=0.05em] \fill [color={rgb,255: red,226; green,194; blue,198}] (0,0) rectangle (0.75em,0.75em); 13,401  &
    \makecell[l]{InternVid-10M} &
    \tikz[baseline=0.05em] \fill [color={rgb,255: red,247; green,230; blue,216}] (0,0) rectangle (0.75em,0.75em); 45,000 &
    \tikz[baseline=0.05em] \fill [color={rgb,255: red,220; green,227; blue,242}] (0,0) rectangle (0.75em,0.75em); 245,840 &
    \tikz[baseline=0.05em] \fill [color={rgb,255: red,169; green,193; blue,217}] (0,0) rectangle (0.75em,0.75em); 48,246 \\
    \makecell[l]{ActivityNet} &
    \tikz[baseline=0.05em] \fill [color={rgb,255: red,222; green,205; blue,180}] (0,0) rectangle (0.75em,0.75em); 7,953 &
    \tikz[baseline=0.05em] \fill [color={rgb,255: red,229; green,240; blue,219}] (0,0) rectangle (0.75em,0.75em); 44,100 &
    \tikz[baseline=0.05em] \fill [color={rgb,255: red,247; green,230; blue,216}] (0,0) rectangle (0.75em,0.75em); 12,771 &
     \makecell[l]{HD-VILA-100M} &
    \tikz[baseline=0.05em] \fill [color={rgb,255: red,182; green,184; blue,214}] (0,0) rectangle (0.75em,0.75em); 48,260 &
    \tikz[baseline=0.05em] \fill [color={rgb,255: red,185; green,215; blue,210}] (0,0) rectangle (0.75em,0.75em); 263,652 &
    \tikz[baseline=0.05em] \fill [color={rgb,255: red,200; green,194; blue,190}] (0,0) rectangle (0.75em,0.75em); 51,743 \\
    \makecell[l]{Kinetics-700} &
    \tikz[baseline=0.05em] \fill [color={rgb,255: red,220; green,227; blue,242}] (0,0) rectangle (0.75em,0.75em); 34,998 &
    \tikz[baseline=0.05em] \fill [color={rgb,255: red,169; green,193; blue,217}] (0,0) rectangle (0.75em,0.75em); 0 &
    \tikz[baseline=0.05em] \fill [color={rgb,255: red,182; green,184; blue,214}] (0,0) rectangle (0.75em,0.75em); 0 &
    \makecell[l]{VIDAL} &
    \tikz[baseline=0.05em] \fill [color={rgb,255: red,216; green,166; blue,166}] (0,0) rectangle (0.75em,0.75em); 55,000 &
    \tikz[baseline=0.05em] \fill [color={rgb,255: red,226; green,194; blue,198}] (0,0) rectangle (0.75em,0.75em); 300,472 &
    \tikz[baseline=0.05em] \fill [color={rgb,255: red,222; green,205; blue,180}] (0,0) rectangle (0.75em,0.75em); 58,968 \\ \bottomrule
\end{tabular}
\end{minipage}
\caption{Distribution of data across different datasets and question types (Caption, Open-ended, and Multi-Choice).}
\label{fig:ov_plot}
\end{figure*}


%
%
\begin{figure*}[t!]
\centering
\includegraphics[width=\textwidth]{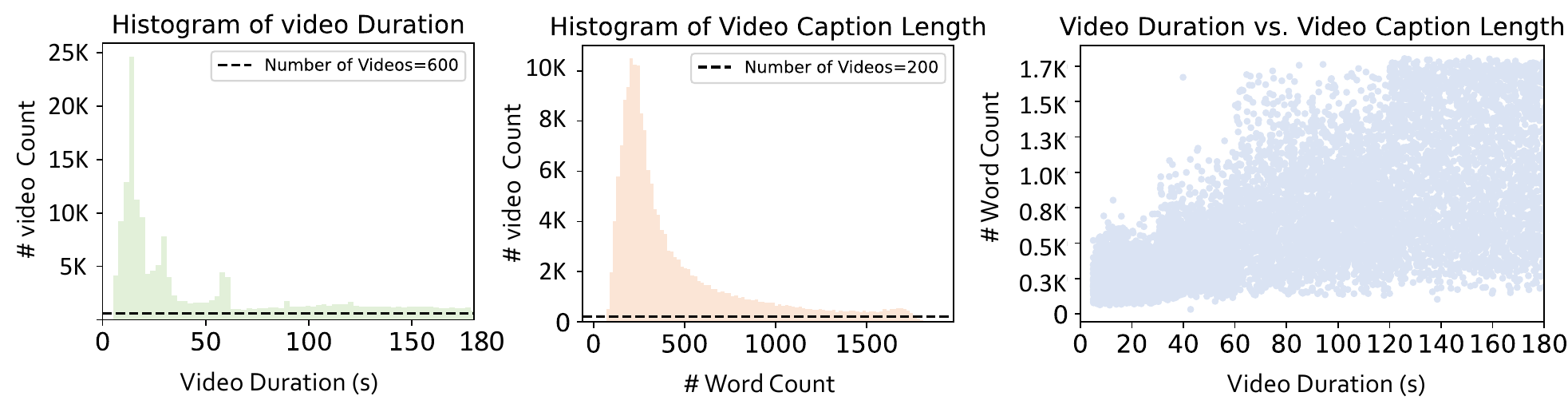}
\caption{ (Left) Visualization of the video duration. (Middle) Visualization of the number of words in the video caption. (Right) Visualization of caption length versus video duration.}
%
\label{fig:dataset_distribution}
\end{figure*}
\begin{figure*}[t!]
\centering
\includegraphics[width=0.95\textwidth]{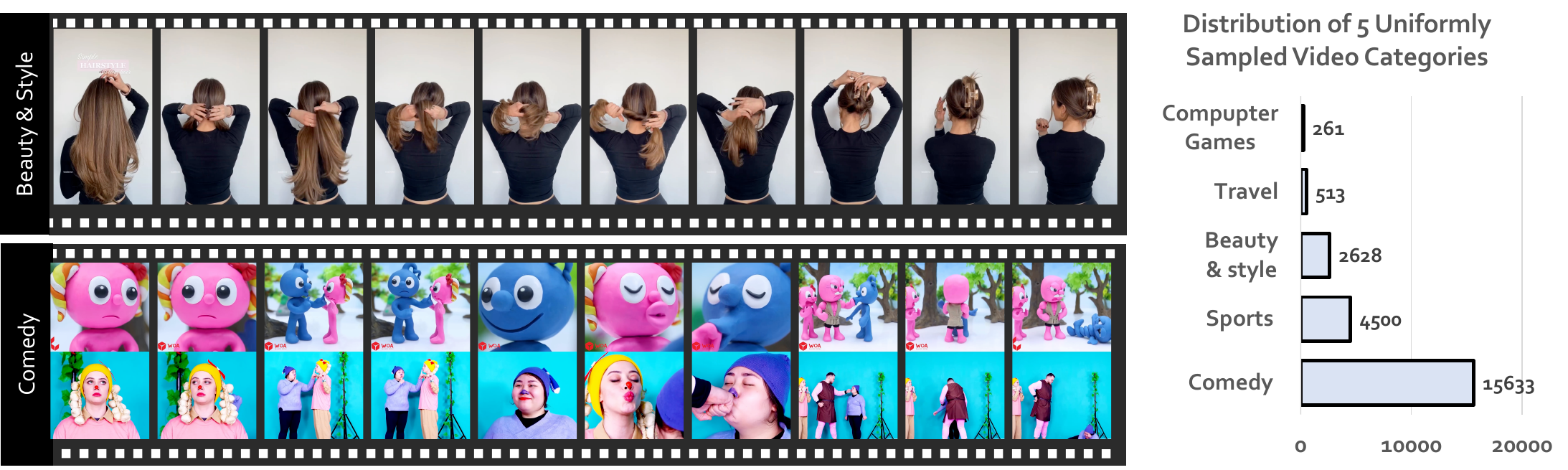}
\caption{(Left) Display of YouTube Shorts across four video categories. (Right) Distribution of 5 uniformly chosen video categories. }
\label{fig:video_category}
\end{figure*}

\begin{table*}[t]
\tabstyle{5pt}
\centering
\small
\caption{\textbf{Comparison of \DataName~and other video-language datasets}. Average FPS represents the average number of frames per second that are used to prompt GPT-4o/GPT-4V for annotation. \ding{72} VIDAL, WebVid, ActivityNet. \ding{110} Panda-70M, Pexels, Pixabay, Mixkit, BDD100K, Ego4d.  \ding{74} HD-VILA-100M, Kinetics-700M, Ego4D, VidOR, InternVid, YouCook2, ActivityNet, Sth-sthv2, VIDAL, Charades.}
\begin{tabular}{l|cccccccc}
\toprule
\multirow{2}{*}{} & \multirow{2}{*}{Text} & Video & \multirow{2}{*}{\#Video} & Total Video  & Average & \multirow{2}{*}{\#Caption} & \#OE & \#MC \\ 
& & Source & & Length &  FPS & &  QA &  QA \\  \midrule
LLaVA-Hound & GPT-4V  & \ding{72} & 900K  & 3Khr              & 0.008                      & 900K & 900K &0               \\
ShareGPT4Video & GPT-4V & \ding{110} & 40K  & 0.2Khr                & 0.15                         & 40K &0 &0               \\
\rowcolor{front-color}
\DataName   & GPT-4o  & \ding{74} & 178K  & 2Khr             & 1                        & 178K& 960K & 196K \\ \bottomrule
\end{tabular}
\label{tab:dataset_comparison}
\end{table*}

\paragraph{Dataset Comparison}
We provide a comparison of high-quality instruction following video-language datasets, with a focus on synthetic data created with strong AI models, as shown in Table~\ref{tab:dataset_comparison}. 
$(i)$   \textit{A broad collection of dynamic videos.} In terms of video sources, although LLaVA-Hound~\citep{zhang2024direct} contains the largest number of videos, 44\% of its video data are sourced from WebVid~\citep{Bain21}, where most videos are static. ShareGPT4Video~\citep{chen2024sharegpt4video} includes 30\% of its videos from Pexels, Pixabay, and Mixkit, which are aesthetically good but also mostly static. Additionally, the majority of its videos come from Panda-70M, which are short clips from longer videos---suggesting simpler plots. In contrast, we carefully select video sources that offer dynamic, untrimmed videos with complex plots, which are crucial for developing a powerful video understanding model.\footnote{Example videos: \href{https://ak.picdn.net/shutterstock/videos/21179416/preview/stock-footage-aerial-shot-winter-forest.mp4}{WebVid},\href{https://pixabay.com/videos/plane-modelling-miniature-lockheed-134519/}{Pixabay},\href{https://www.pexels.com/video/a-bird-is-standing-on-the-beach-27916646/}{Pexels},\href{https://mixkit.co/free-stock-video/a-young-woman-clad-in-snugly-black-sportswear-doing-lunges-52112/}{Mixkit}.}
$(ii)$ \textit{High frames per second}. Regarding frame sampling in language annotations, the proposed datasest considers 1 FPS, while other datasets consider much lower FPS.  LLaVA-Hound uniformly samples 10 frames from videos of any length. The average FPS is 0.008, which may miss some fine details. ShareGPT4Video picks key frames using CLIP~\citep{radford2021learning} based on frame uniqueness. This method might also miss subtle changes in the video because CLIP embeddings do not capture fine-grained dynamics well. Our method samples FPS=1 without using key frame selection algorithms, ensuring the detailed temproal information can be expressed in annotations and high coverage. 
$(iii)$   \textit{Diverse tasks.} The proposed dataset considers three common task types, including caption, free-form and closed-form QA, while existing datasets only consider a subset. Meanwhile, the quality and numbers of samples in our dataset is higher.



\section{Experiments}
We conducted evaluations for the LLaVA-Video models across all benchmarks using LMMs-Eval~\citep{zhang2024lmms} to ensure standardization and reproducibility. To fairly compare with other leading video LMMs, we primarily used results from original papers. When results were not available, we integrated the models into LMMs-Eval and assessed them under consistent settings. Following LLaVA-OneVision~\citep{li2024llavaonevision}, we employed SigLIP~\citep{zhai2023sigmoid} as our vision encoder, and Qwen2~\citep{yang2024qwen2} as the LLM. The LLaVA-Video model builds on the single-image (SI) stage checkpoint from the LLaVA-OneVision model~\citep{li2024llavaonevision}, which was trained using only image data. 

 %

\paragraph{Video Representations}
Following the classic SlowFast idea in video representations~\citep{feichtenhofer2019slowfast,xu2024slowfast,huang2024lita}, we develop \textit{\MethodName{}} to optimize the balance between the number of frames and the count of visual tokens, within the budget of the limited context window in LLM and GPU memory for video representation. Please refer to Appendix~7 for detailed information. Specifically, we represent each video as a sequence with maximum $T$ frames. Each frame is represented in $M$ tokens. we categorize the frames into two groups, based on the a strike rate $s$, where the every $s$ frames are uniformly selected to form the \textit{slow} frame group, and the rest of the frames are consdiered as the \textit{fast} frame group. Note that a special case $s=1$ leads to only one group, reducing the SlowFast representation to the original simple representation. For each group, we apply different pooling rate using Pytorch function pooling $\mathtt{avg\_pool2d}()$.  $p \times p$ pooling and $2p \times 2p$ pooling for slow and fast frames, respectively. To summarize, we paramterize the video representation configuration as $\Vcal = (T, M, s, p)$. The total number of tokens is \(
\#tokens = \left\lfloor T/s \right\rfloor \times \left\lfloor M/p^2 \right\rfloor + \left( T - \left\lfloor T/s \right\rfloor \right) \times \left\lfloor M/4p^2 \right\rfloor
\)

\paragraph{Evaluation Benchmarks.} For full evaluation, we consdier 11 video benchmarks. conducted tests across various video captioning , video open-ended question-answering and video multiple-choice question-answering benchmarks, including ActivityNet-QA~\citep{yu2019activityqa}, which features human-annotated action-related QA pairs from the ActivityNet dataset. We also utilized LongVideoBench~\citep{wu2024longvideobench}, EgoSchema \citep{mangalam2024egoschema}, and MLVU~\citep{zhou2024mlvu} for long video understanding, PerceptionTest~\citep{patraucean2023perception} for assessing fine-grained perception skills, and VideoMME~\citep{fu2024videomme} and NExT-QA~\citep{xiao2021next} for diverse video domains and durations. Additional tests included VideoDetailCaption~\citep{videodetail2024}, Dream-1K~\citep{wang2024tarsierrecipestrainingevaluating}, Video-ChatGPT \citep{Maaz2023VideoChatGPT} for detailed video descriptions, TemporalBench~\cite{cai2024temporalbench} for fine-grained temporal understanding.

For ablation studies in .~\ref{sec;subsec:dataset_ablation} and Sec.~\ref{sec;subsec:dataset_comparison}, we conduct evaluation across 4 datasets. NExT-QA~\citep{xiao2021next} and PerceptionTest~\citep{patraucean2023perception}, which use training data from the \DataName{}, are treated as in-domain datasets. Conversely, VideoMME~\citep{fu2024videomme} and EgoSchema~\citep{mangalam2024egoschema} are consider as zero-shot datasets.

 \subsection{Overall Results}

We fine-tune LLaVA-OneVision (SI) on the joint dataset of video and image data. Specifically, we added video data from the \DataName{} dataset and four public datasets: 
ActivityNet-QA~\citep{yu2019activityqa}, NExT-QA~\citep{xiao2021next}, PerceptionTest~\citep{patraucean2023perception}, and LLaVA-Hound-255K~\citep{zhang2024direct}, focusing on videos shorter than three minutes. These datasets were selected to improve our model’s performance, contributing to a total of 1.6 million video-language samples, which include 193,510 video descriptions, 1,241,412 open-ended questions, and 215,625 multiple-choice questions. Remarkably, 92.2\% of the video descriptions, 77.4\% of the open-ended questions, and 90.9\% of the multiple-choice questions were newly annotated. Additionally, we used 1.1 million image-language pairs from the LLaVA-OneVision model \citep{li2024llavaonevision}. 
%
We consider the same video representation configurations for the training and inference stages. On 128 NVIDIA H100 GPUs, the video representations for LLaVA-Video-7B and LLaVA-Video-72B are $\Vcal = (64, 679, 1, 2)$ and $\Vcal = (64, 679, 3, 2)$, respectively.

\begin{table*}[t!]
\tabstyle{1pt}
\centering
\small
\caption{\ModelName{} performance on video benchmarks. We report the score out of 5 for VideoDC, VideoChatGPT while other results are reported in accuracy. All results are reported as 0-shot accuracy. *indicates that the training set has been observed in our data mixture.}
\begin{tabular}{@{}lcc|cc|ccccccccc@{}}
    \toprule
     & \multicolumn{2}{c}{\scriptsize{Caption}} & \multicolumn{2}{c}{\scriptsize{Open-Ended Q\&A}} & \multicolumn{7}{c}{\scriptsize{Multi-Choice Q\&A}}  \\   
    \multirow{2}{*}{\textbf{Model}} & \rotatebox{90}{\textbf{\scriptsize{VideoDC}}} & \rotatebox{90}{\textbf{\scriptsize{Dream-1K}}} & \rotatebox{90}
    {\textbf{\scriptsize{ActNet-QA}}} & \rotatebox{90}{\textbf{\scriptsize{VideoChatGPT}}} & \rotatebox{90}{\textbf{\scriptsize{EgoSchema}}} & \rotatebox{90}{\textbf{\scriptsize{MLVU}}} & \rotatebox{90}{\textbf{\scriptsize{MVBench}}} & \rotatebox{90}{\textbf{\scriptsize{NExT-QA}}} & \rotatebox{90}{\textbf{\scriptsize{PerceptionTest}}}  & \rotatebox{90}{\textbf{\scriptsize{LongVideoBench}}} &\rotatebox{90} 
    {\textbf{\scriptsize{TemporalBench}}} &\rotatebox{90} 
    {\textbf{\scriptsize{VideoMME}}} \\ \cmidrule(l){2-13} 
    & test & test & test & test & test & m-avg & test & mc & val   & val & m-acc & wo/w-subs \\ \midrule
    \multicolumn{11}{l}{\textit{Proprietary models}} \\
    \rowcolor{Gray}
    GPT-4o~\citep{openai2024gpt4o} & - & 39.2 & - & - & - & 64.6 & - & - & -    & 66.7 & 35.3 & 71.9/77.2 \\
    \rowcolor{Gray}
    Gemini-1.5-Pro~\citep{team2023gemini} & -  & 36.2 & 57.5 & - & 72.2 & - & -  & - & -  & 64.0 & 25.6 & 75.0/81.3 \\ \midrule 
    \multicolumn{11}{l}{\textit{Open-source models}} \\
    VILA-40B~\citep{lin2024vila} & 3.37 & 33.2 & 58.0 & 3.36 & 58.0 & - & - & 67.9 & 54.0  & - & - & 60.1/61.1 \\
    PLLaVA-34B~\citep{xu2024pllava} & -  & 28.2 & 60.9 & 3.48  & - & - & 58.1  & - & - & 53.2 & - & - \\    
    LongVA-7B~\citep{zhang2024long} & 3.14 & - & 50.0 & 3.20 & - & 56.3 & - & 68.3 & -  & -  & - & 52.6/54.3 \\
    IXC-2.5-7B~\citep{zhang2024internlm} & - & - & 52.8 & 3.46 & - & 37.3 & 69.1 & 71.0 & 34.4    & - & 16.7 & 55.8/58.8 \\   
    LLaVA-OV-7B~\citep{li2024llavaonevision} & 3.75 & 31.7 & 56.6 & 3.51 & 60.1 & 64.7 & 56.7 & 79.4* & 57.1    & 56.5 & 18.7 & 58.2/61.5 \\
    VideoLLaMA2-72B~\citep{cheng2024videollama2advancingspatialtemporal} & - & 27.1 & 55.2 & 3.16  & 63.9 & 61.2 & 62.0 & - & -  & - & - & 61.4/63.1 \\ 
    LLaVA-OV-72B~\citep{li2024llavaonevision} & 3.60 & 33.2 & 62.3 & 3.62  & 62.0 & 68.0 & 59.4 & 80.2* & 66.9  & 61.3 & 26.6 & 66.2/69.5 \\ \midrule 
    \rowcolor{front-color}
    \ModelName{}-7B & 3.66 & 32.5 & 56.5* & 3.52 & 57.3  & 70.8 & 58.6 & 83.2* & 67.9*   & 58.2 & 22.9 & 63.3/69.7 \\ 
    \rowcolor{front-color}
    \ModelName{}-72B & 3.73 & 34.0 & 63.4*  & 3.62 & 65.6 & 74.4 & 64.1 & 85.4* & 74.3*  & 61.9 & 33.7 & 70.5/76.9 \\ 
    \bottomrule
    \end{tabular}%
\label{tab:video-bench}
\end{table*}

In Table~\ref{tab:video-bench}, we compare the performance of different models on various video benchmarks. The 72B model performs as well as the commercial, closed-source model Gemini-1.5-Flash~\citep{team2023gemini}, highlighting the effectiveness of open-source efforts in achieving comparable results. The \ModelName{}-7B model outperforms the previous top model, LLaVA-OV-7B, in seven out of ten datasets. Analysis of individual datasets shows some noteworthy trends. For instance, on benchmarks like MLVU, LongVideoBench, and VideoMME, which primarily use video data from YouTube, this improvement may be due to the inclusion of extensive YouTube data in \DataName{}, as illustrated in Fig.~\ref{fig:ov_plot}. Additionally, the improvement on ActivityNet-QA is small; this could be because many questions in ActivityNet-QA, such as ``What’s the color of the ball?'' can be answered by viewing a single frame. The visibility of the ball from the beginning to the end of the video means understanding the video sequence is unnecessary, so \DataName{} offers little advantage in this context.
We find that \ModelName{}-7B is notably weaker in the specialized task of EgoSchema, an ego-centric dataset. This weakness may be due to a significant reduction in the proportion of ego-centric data in the training dataset of \ModelName{}. However, this impact is less pronounced in larger models, as demonstrated by the \ModelName{}-72B model's superior performance over LLaVA-OV-72B in EgoSchema.



\subsection{Dataset Ablation}
\label{sec;subsec:dataset_ablation}
Note that the training set for \ModelName{} includes six datasets: \DataName{}, LLaVA-Hound~\citep{zhang2024direct}, NExT-QA~\citep{xiao2021next}, ActivityNet-QA~\citep{yu2019activityqa}, PerceptionTest~\citep{patraucean2023perception}, and image data from LLaVA-OneVision~\citep{li2024llavaonevision}. In this section, we conduct ablation studies to assess the impact of each dataset. We separately fine-tune the LLaVA-OneVision (SI) model for each experimental setting, progressively adding datasets to the baseline.

The results are presented in Table~\ref{tab:dataset_ablation}. Initially, we used a basic model trained solely on the LLaVA-Hound dataset as our baseline. Compared to this baseline, adding the \DataName{} dataset significantly improved performance, enhancing scores in both in-domain and out-of-domain tasks. Specifically, we observed a 31.9-point increase in NExT-QA scores and a 9.1-point rise in VideoMME scores. Furthermore, including the PerceptionTest dataset enhanced its associated task. Additionally, integrating high-quality image data provided modest benefits on EgoSchema.

\begin{table}[t!]
\tabstyle{5pt}
\centering
\small
\caption{Ablation study on the \ModelName{} model with various configurations of training data. Three Q\&A datasets indicate: NExT-QA, ActivityNet-QA and PerceptionTest.}
\begin{tabular}{@{}l|llll@{}}
    \toprule
        & \multicolumn{2}{c}{in-domain} & \multicolumn{2}{c}{out-of-domain} \\   

      \multirow{2}{*}{Method}  &
    \rotatebox{90}{\textbf{\scriptsize{NExT-QA}}} & \rotatebox{90}{\textbf{\scriptsize{PercepTest}}} & 
    \rotatebox{90}{\textbf{\scriptsize{EgoSchema}}} & 
    \rotatebox{90}{\textbf{\scriptsize{VideoMME}}} \\ \cmidrule(l){2-5} 
      &    mc & val  & test & wo \\ \midrule
     LLaVA-Hound  & 64.4 & 51.4 & 51.0 & 54.1 \\
     +\DataName{} & 80.1  & 57.1  & 56.5   & 63.2   \\ 
     
     +Three Q\&A datasets & 80.1 & 69.0   & 55.6  & 61.9   \\

     +LLaVA-OV (images) & 83.2  & 67.9  & 57.3   & 63.4   \\
    \bottomrule
\end{tabular}

\label{tab:dataset_ablation}
\end{table}

\subsection{Dataset Comparison}
\label{sec;subsec:dataset_comparison}
We conduct two ablation studies to analyze our dataset and training strategy. In Table~\ref{tab:dataset_comparison_exp}, we compared three datasets where the language annotations are from GPT-4V/GPT-4o. For each experiment, we fine-tune the LLaVA-OneVision (SI) model separately on each specific dataset setting.

Two group of experiments are considered to assess the data quality of \DataName{} compare to LLaVA-Hound and ShareGPT4Video. In the first group, to compare \DataName{} with LLaVA-Hound, we randomly selected 900K open-ended questions to match the number in LLaVA-Hound. We included all captions and did not sample the multiple-choice questions. In the second group, comparing \DataName{} to ShareGPT4Video, we randomly sampled 40K video captions to align with those in ShareGPT4Video. Since ShareGPT4Video lacks open-ended and multiple-choice questions, we supplemented with annotations from NExT-QA, PerceptionTest, and ActivityNet-QA. 
In the first group of Table~\ref{tab:dataset_comparison_exp}, we compare \DataName{} with LLaVA-Hound. Although LLaVA-Hound has more captions than \DataName{}, our results are still better. The quality of LLaVA-Hound is limited due to two main issues: (1) Static video: Its primary video source is WebVid~\citep{Bain21}, which tends to have relatively static content. (2) Sparse sampling: its sampling rate of 10 frames per video leads to annotations that do not fully capture the complete plot of the video. This underscores that the quality of video instruction-following data is more important than its quantity. Additionally, the second experiment group in Table~\ref{tab:dataset_comparison_exp} shows that the model trained with \DataName{} outperforms that of ShareGPT4Video,  highlighting the superiority of our data's quality.


\begin{table}[t!]
\tabstyle{1pt}
\centering
\small
\caption{Comparison of \DataName{} and other video instruction-following datasets.}
\begin{tabular}{@{}llll|llll}
    \toprule
       & & & & \multicolumn{2}{c}{in-domain} & \multicolumn{2}{c}{out-of-domain} \\   
     &
      \multirow{3}{*}{\textbf{\#Cap}} &
    \multirow{3}{*}{\textbf{\#OE}} &
    \multirow{3}{*}{\textbf{\#MC}} &
    \rotatebox{90}{\textbf{\scriptsize{NExT-QA}}} & \rotatebox{90}{\textbf{\scriptsize{PercepTest}}} & 
    \rotatebox{90}{\textbf{\scriptsize{EgoSchema}}} & 
    \rotatebox{90}{\textbf{\scriptsize{VideoMME}}}  \\ \cmidrule(l){5-8} 
      & &  &   & mc & val  & test & wo \\ \midrule
    LLaVA-Hound & 900K & 900k & 0 & 39.8 & 53.1 & 25.8 & 55.2 \\  
    LLaVA-V-178K & 178K & 900k & 0 & 73.2  & 55.9 & 49.8  & 59.6  \\
  \midrule
    ShareGPT4Video & 40K & 40K & 19K & 69.6 & 55.2 & 58.9 & 51.0 \\  
    LLaVA-V-178K  & 40K & 40K & 19K & 75.8  & 55.4 & 55.8  & 53.5  \\
    \bottomrule
    \end{tabular}%
\label{tab:dataset_comparison_exp}
\end{table}


\section{Conclusion}
This study introduces the \DataName{} dataset, a high-quality synthetic dataset for video-language instruction-following. It is favored for its dense frame sampling rate in longer, untrimmed videos, covering diverse tasks such as captioning, open-ended and multi-choice QA. By training on the joint dataset of \DataName{} with existing visual instruction tuning data, we developed a new model family, \ModelName{}, which considers video representation to effectively use GPU resources. This allows us to include more frames in the training process. The experimental results have demonstrated the effectiveness of the proposed synthetic dataset, and \ModelName{} models have achieved excellent performance on a wide range of video benchmarks.

\section{Limitations}
The videos in \DataName{} are sourced from various platforms. This diversity introduces potential biases inherent in these sources. Furthermore, there is a concern regarding the potential skew in the question-answer pairs, possibly influenced by the annotators' perspectives. 

\section*{Acknowledgement}
This study is supported by the Ministry of Education, Singapore, under its MOE AcRF Tier 2 (MOE-T2EP20221-0012, MOE-T2EP20223-0002), and under the RIE2020 Industry Alignment Fund – Industry Collaboration Projects (IAF-ICP) Funding Initiative, as well as cash and in-kind contribution from the industry partner(s).
\clearpage

\clearpage
\bibliography{main}
\bibliographystyle{tmlr}

\clearpage
\appendix








\section{Video Representations}
\label{app:visual_representations}
\subsection{Efficient Video Representations in LMMs}
Current designs of large multimodal models (LMM) typically connect a vision encoder~\citep{radford2021learning,zhai2023sigmoid} to a large language model~\citep{yang2024qwen2} through a lightweight projector~\citep{liu2024visual} or a resampler~\citep{li2023blip2bootstrappinglanguageimagepretraining,alayrac2022flamingovisuallanguagemodel}. These components transform a set of visual representations into ``visual tokens'' aligned with text embeddings. In contrast to image-based LMMs, which generate only a small number of visual tokens easily managed by a standard GPU, video LMMs face challenges due to a large number of visual tokens derived from multiple video frames. The LLaVA-NeXT-Video~\citep{zhang2024llavanext-video} and PLLaVA~\citep{xu2024pllava} models address this by simly considering average pooling to reduce the number of tokens representing each frame. 

Following the idea of SlowFast in the traditional video understanding~\citep{feichtenhofer2019slowfast}, adaptive reductions in visual tokens are demonstrated by recent video LMMs, LITA~\citep{huang2024lita} and SlowFast-LLaVA~\citep{xu2024slowfastllavastrongtrainingfreebaseline}. Initially, these methods represent all sampled frames with a minimal number of visual tokens (fast frame)--- typically just one---by using a large pooling stride. They then switch to a smaller pooling stride for certain frames to retain more visual tokens (slow frame). Finally, they combine the visual tokens of fast frames with those of slow frames. However, this approach can lead to some frames being represented twice. In contrast, our method uses a larger pooling stride for sampled frames to maintain fewer visual tokens (fast frame) \textit{or} a smaller stride for others to keep more (slow frame). We then arrange slow and fast frames in an interleaving pattern.

\begin{figure*}[h]
\centering
\includegraphics[width=0.95\textwidth]{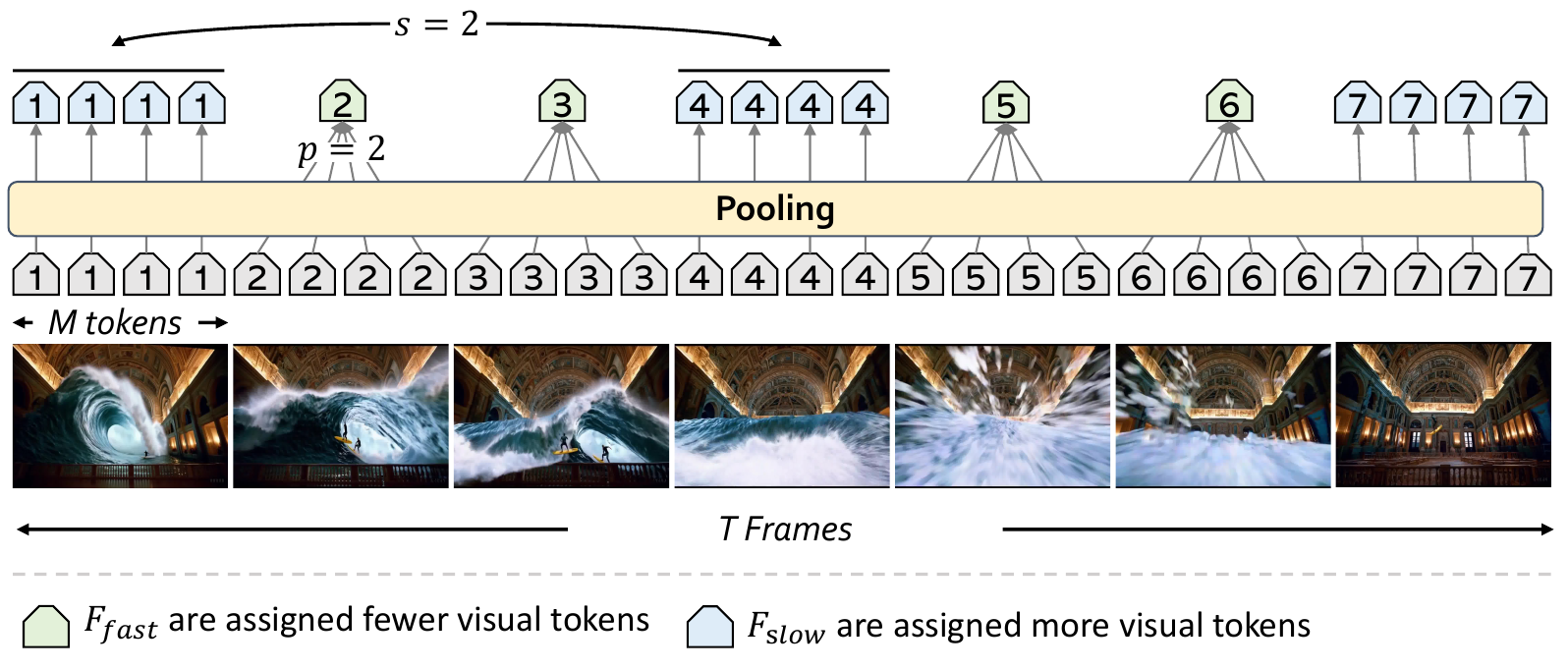}
\caption{Video representations. A different number of tokens are utilized to represent frames.}
\label{fig:llava_arch}
\end{figure*}


\subsection{\MethodName}
We represent each video as a sequence with maximum $T$ frames. Each frame is represented in $M$ tokens. FPS-based video representation can be considered in the future. 
Specifically, each frame is encoded via an image encoder and a two-layer MLP for projection. These visual tokens are concatenated with word tokens and processed by a large language model (LLM). 
Managing tokens for every frame can be computationally demanding. 
For instance, employing the SigLIP~\citep{zhai2023sigmoid} encoder for a video with \(T = 100\) results in 67,600 tokens, assuming \(M = 729\) tokens per frame, which often exceeds GPU memory limits. This issue is exacerbated when using large-parameter LLMs; with the Qwen2-72B model, we could only process 8 frames before maxing out the memory on 128 NVIDIA H100 GPUs. Such a limited number of frames can introduce inconsistencies in language annotations, reducing model efficacy. 
One strategy to incorporate more frames is by applying \(p \times p\) spatial average pooling to reduce \(M\) to \(M/p^2\), thus lowering the token count per frame as suggested by recent studies~\citep{xu2024pllava,zhang2024llavanext-video}. However, the number of visual tokens is crucial for preserving the informational content of each frame, which is vital for video comprehension.

In our \MethodName, we categorize the frames into two groups, based on the a strike rate $s$, where the every $s$ frames are uniformly selected to form the \textit{slow} frame group, and the rest of the frames are consdiered as the \textit{fast} frame group. Note that a special case $s=1$ leads to only one group, reducing the SlowFast representation to the original simple representation. For each group, we apply different pooling rate using Pytorch function pooling $\mathtt{avg\_pool2d}()$.  $p \times p$ pooling and $2p \times 2p$ pooling for slow and fast frames, respectively. To summarize, we paramterize the video representation configuration as $\Vcal = (T, M, s, p)$. The total number of tokens is \(
\#tokens = \left\lfloor T/s \right\rfloor \times \left\lfloor M/p^2 \right\rfloor + \left( T - \left\lfloor T/s \right\rfloor \right) \times \left\lfloor M/4p^2 \right\rfloor
\)

\section{Data}
\subsection{Video Detail Description}
\begin{figure*}[h]
\centering
\includegraphics[width=0.95\textwidth]{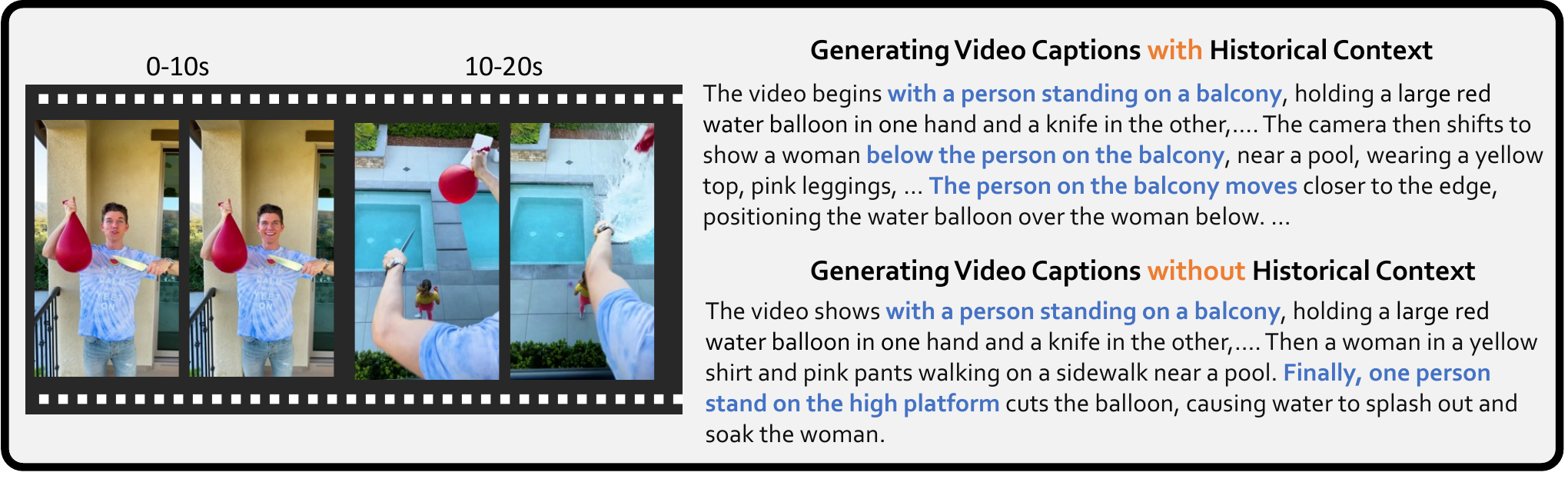}
\caption{Generating video captions with or without historical context.}
\label{fig:in_context}
\end{figure*}

As discussed in Section~3.2, we show that generating \textit{level-1 description} should consider historical context. Figure~\ref{fig:in_context} illustrates the impact of excluding historical context on the quality of video descriptions. Specifically, including historical context helps accurately identify characters across different times as the same individual.

\begin{table*}[h!]\centering
\caption{Question types for video question answering in data creation. For each type, we provide its name, description, and the proportion it represents in the \DataName.}
\tabstyle{3pt}
\begin{minipage}{0.95\textwidth}\vspace{0mm}    
\centering
\begin{tcolorbox} 
    \centering
   
      \footnotesize
\begin{tabular}{>{\raggedright\arraybackslash}m{0.25\linewidth}|>{\centering\arraybackslash}m{0.6\linewidth}|>{\centering\arraybackslash}m{0.1\linewidth}}
\toprule
\textbf{Question type} & \textbf{Description} & \textbf{Proportion}\\
\midrule
Temporal & Designed to assess reasoning about temporal relationships between actions/events. Questions involve previous, present, or next actions. & 7.2\% \\ \midrule
Spatial & Tests ability to perceive spatial relationships between observed instances in a video scene.  & 7.2\% \\ \midrule
Causal & Focuses on explaining actions/events, determining intentions of actions or causes for subsequent events.  & 7.2\%\\ \midrule
Description-Scene & Assesses ability to describe the major scene of the video, like where it takes place and the overall environment.  & 7.2\% \\ \midrule
Description-Human & Involves describing actions or attributes of people, such as their activities and appearances.  & 6.7\% \\ \midrule
Description-Object & Assesses ability to describe attributes of objects, like their appearance and function.  & 7.0\% \\ \midrule
Count & Tests ability to count instances of objects, people, actions, and to distinguish between old and new elements in a scene.  & 7.1\%\\ \midrule
Binary & Involves yes or no questions related to the video content.  & 7.2\%\\ \midrule
Fine Grained Action Understanding & Creates questions challenging comprehension of subtle actions.  & 6.5\%\\ \midrule
Plot Understanding & Challenges ability to interpret the plot in the video.  & 7.1\% \\ \midrule
Non-Existent Actions with Existent Scene Depictions & Assesses reasoning with introduced non-exist ent activities without changing physical details.  & 6.6\%\\ \midrule
Time Order Understanding & Challenges recognition of temporal sequence of activities in videos.  & 6.9\% \\ \midrule
Object Direction & Emphasizes perception of object movement direction.  & 3.8\% \\ \midrule
Camera Direction & Focuses on the direction of camera movement.  & 4.1\% \\ \midrule
Speed & Delves into discerning variations in speed, including absolute and relative speeds.  & 3.6\% \\ \midrule
Attribute Change & Centers on how attributes of objects or the entire video change over time, like size, shape, color, and more.  & 4.5\% \\\bottomrule
\end{tabular}
\end{tcolorbox}
\label{tab:question_dimension}
\end{minipage}
\end{table*}

\begin{table*}[h!]
\centering
\begin{minipage}{0.99\textwidth}
\vspace{0mm}    
\centering
\begin{tcolorbox} 
    \centering
    \small
    \hspace{-6mm}
    \begin{tabular}{p{0.99\textwidth}}
    \begin{minipage}{0.99\textwidth}
    \vspace{0mm}

    \VarSty{tasks} = ``\\
    \# Temporal: this task is designed to assess the capability of reasoning ...$<$omitted$>$ \\
    \#\# caption-1: The video features a child sitting in a baby chair at a dining table, creating...$<$omitted$>$ \\
    \#\# question-1: What was the child doing as he sat on the baby chair? \\
    \#\# answer-1: The child was reading a book. \\
    ... \\
    \#\# caption-3: ...$<$omitted$>$ \\
    \#\# question-3: ...$<$omitted$>$\\
    \#\# answer-3: ...$<$omitted$>$ \\
    \# Spatial: this task involves creating questions that test a person's ability...$<$omitted$>$ \\
    ...$<$omitted$>$ '' \\
    \VarSty{system\_message} = ``\\
    \#\#\# Task: \\
    Given a detailed description that summarizes the content of a video, generate question-answer pairs based on the description to help humans better understand the video.
    The question-answer pairs should be faithful to the content of the video description and developed from different dimensions to promote comprehensive understanding of the video. \\
    Here are some question dimensions and their explanations and exampled question-answer pairs for reference: \\
    \textit{\{task\_definitions\}} \\
    \#\#\#\# Guidelines For Question-Answer Pairs Generation: \\
    - Read the video description provided carefully, paying attention to the content, such as the scene where the video takes place, the main characters and their behaviors, and the development of the events. \\
    - Generate appropriate question-answer pairs based on the description. The question-answer pairs should cover as many question dimensions and not deviate from the content of the video description. \\
    - Generate 1 question-answer pair for each dimension.
    \\
    \#\#\# Output Format: \\
    1. Your output should be formed in a JSON file. \\
    2. Only provide the Python dictionary string. \\
    Your response should look like: \\
    \texttt{[{{"Dimension": <dimension-1>, "Question": <question-1>, "Answer": <answer-1>}},} \\
    \texttt{{{"Dimension": <dimension-2>, "Question": <question-2>, "Answer": <answer-2>}}...]} '' \\
    \VarSty{user\_message} = `` \\
    Please generate question-answer pairs for the following video description: \\
    Description: \{caption\} '' \\
    \\
    \For{\VarSty{cur\_video} in \VarSty{videos}}{ 
        \var{sys\_msg = \VarSty{system\_messages}.format(task\_definitions=\VarSty{tasks})} \\
         \var{usr\_msg = \VarSty{user\_messages}.format(caption=\VarSty{cur\_video}) \\
         response = GPT4O(sys\_msg,usr\_msg)}
    } 
    \end{minipage}
    \end{tabular}
\end{tcolorbox}
    
\caption{We explain the process of creating prompts for GPT-4O to gather question-answer pairs from each video description. \VarSty{tasks} includes the definition of all question types along with examples of question-answer pairs. We instruct GPT-4O to generate questions that cover as many question types as possible.}
\label{tab:prompt_conversation}
\end{minipage}
\end{table*}

\subsection{Video Question Answering}
In Table~\ref{tab:question_dimension}, we list the names and descriptions of different question types and their corresponding proportions in the \DataName{} dataset. The prompt used to generate video question-answer pairs from GPT-4O is shown in Table.~\ref{tab:prompt_conversation}. In Fig.~\ref{fig:annotation_example}, we show an example of a video along with its detailed description, an open-ended question, and a multiple-choice question.

\subsection{Dataset Comparison}

\begin{table*}[h!]
\tabstyle{1pt}
\centering
\small
\caption{\textbf{Comparison of \DataName~and other video-language datasets}. Average FPS represents the average number of frames per second that are used to prompt GPT-4o/GPT-4V for annotation.}
\begin{tabular}{l|ccccccc}
\toprule
\multirow{2}{*}{} & \multirow{2}{*}{Text}  & \multirow{2}{*}{\#Video} & Total Video  & Average & \multirow{2}{*}{\#Caption} & \#OE & \#MC \\ 
&  & & Length &  FPS & &  QA &  QA \\  \midrule
HowTo100M~\citep{miech19howto100m} & ASR & 136M & 134.5Khr & - & 136M&0&0 \\ 
ACAV~\citep{lee2021acav100m} & ASR & 100M & 277.7Khr & - & 100M&0&0 \\ 
YT-Temporal-180M~\citep{zellers2021merlot} & ASR & 180M & - & - & 180M&0&0 \\ 
HD-VILA-100M~\citep{xue2022hdvila} & ASR & 103M & 371.5Khr & - & 103M&0&0 \\ \midrule
MSVD~\citep{chen2011msvd} & Manual & 1970 & 5.3h & - & 1K&0&0 \\ 
LSMDC~\citep{rohrbach2015LSMDC} & Manual & 118K & 158h & - & 118K&0&0 \\ 
MSR-VTT~\citep{xu2016msrvtt} & Manual & 10K & 40h & - & 10K&0&0 \\ 
DiDeMo~\citep{anne2017dimemo} & Manual & 27K & 87h & - & 27K&0&0 \\ 
ActivityNet~\citep{caba2015activitynet} & Manual & 100K & 849h & - & 100K&0&0 \\ 
YouCook2~\citep{Zhou2017YouCookIID} & Manual & 14K & 176h & - & 14K&0&0 \\ 
TVQA~\citep{lei2018tvqa} & Manual & 21K & 3.39Khr & - & 0&0&152K \\ 
ActivityNet-QA~\citep{yu2019activityqa} & Manual & 5.8K & 290h & - & 0&58K&0 \\ 
Social-IQ~\citep{zadeh2019social} & Manual & 1.2K & 20h & - & 0&0&7.5k \\ 
NExT-QA~\citep{xiao2021next} & Manual & 5.4K & 66h & - & 0&52K&47K \\ 
\midrule
MSVD-QA~\citep{xu2017msvdqa} & Open-source Model &  1.9K & 5.3h & - & 41K&50K&0 \\ 
MSRVTT-QA~\citep{xu2017msvdqa} & Open-source Model & 10K & 40h & - & 0 &243K&0 \\ 
Panda-70M~\citep{chen2024panda70m} & Open-source Model & 70.8M & 166.8Khr & - & 70.8M & 0 &0 \\ \midrule
LLaVA-Hound~\citep{zhang2024direct} & GPT-4V  & 900K  & 3Khr              & 0.008                      & 900K & 900K &0               \\
ShareGPT4Video~\citep{chen2024sharegpt4video} & GPT-4V  & 40K  & 0.2Khr                & 0.15                         & 40K &0 &0               \\
\rowcolor{front-color}
\DataName   & GPT-4o  & 178K  & 2Khr             & 1                        & 178K& 960K & 196K \\ \bottomrule
\end{tabular}
\label{tab:dataset_comparison_complete}
\end{table*}

We provide a more comprehensive comparison of \DataName{} with other video-language datasets for the video caption task and video question answer task. Specifically, we organize the table into four groups, each characterized by its method of text annotation.  As shown in Table~\ref{tab:dataset_comparison_complete}, unlike other datasets, \DataName{} uniquely includes all three types of annotations: captions, open-ended questions, and multiple-choice questions.

\begin{table*}[t]
\tabstyle{6pt}
\centering
\small
\caption{Visual Representation Configurations and Performance Correlation. \(T^\text{train}\) and \(T^\text{test}\) are the number of frames in the training and inference stage, respectively. \(M/{p^2}\): number of visual tokens per frame.}
\begin{tabular}{@{}lll|llll@{}}
    \toprule
       & & & \multicolumn{2}{c}{in-domain} & \multicolumn{2}{c}{out-of-domain} \\   

      \multirow{3}{*}{\(T^\text{train}\)} &
    \multirow{3}{*}{\(T^\text{test}\)} &
    \multirow{3}{*}{\(M/p^2\)} &
    \textbf{NExT-QA} & \textbf{PerceptionTest} & 
    \textbf{EgoSchema} & 
    \textbf{VideoMME} \\ \cmidrule(l){4-7} 
      &  &   & mc & val  & test & wo \\ \midrule
  \multicolumn{7}{l}{\textit{Training with more frames} } \\
    32 & 32 & 169 & 80.4 & 68.2 & 56.3 & 59.1 \\ 
    64 & 64 & 169 & 81.4 \textcolor{green}{\scriptsize (+1.0)} & 68.3 \textcolor{green}{\scriptsize (+0.1)} & 58.4 \textcolor{green}{\scriptsize (+2.1)} & 59.6 \textcolor{green}{\scriptsize (+0.5)} \\
  110 & 110 & 169 &  82.0  \textcolor{green}{\scriptsize (+1.6)} & 68.3 \textcolor{green}{\scriptsize (+0.1)} & 59.1 \textcolor{green}{\scriptsize (+2.8)}& 60.4 \textcolor{green}{\scriptsize (+1.3)} \\
  \midrule
  \multicolumn{7}{l}{\textit{Inference with more frames} } \\
  32 & 32 & 169 &  80.4 & 68.2 & 56.3& 59.1 \\
  32 & 64 & 169 &  80.7 \textcolor{green}{\scriptsize (+0.3)} & 68.9 \textcolor{green}{\scriptsize (+0.7)} &  56.3 \textcolor{green}{\scriptsize (+0.0)} &  59.9 \textcolor{green}{\scriptsize (+0.8)}  \\ 
  32 & 110 & 169 & 80.5 \textcolor{green}{\scriptsize (+0.1)} & 67.2 \textcolor{red}{\scriptsize (-1.0)} & 55.2 \textcolor{red}{\scriptsize (-1.1)} & 58.8 \textcolor{red}{\scriptsize (-0.3)} \\
    \midrule
  \multicolumn{7}{l}{\textit{Using more frames with fewer visual tokens per frame} } \\
  32 & 32 & 729 & 79.4 &  69.5 & 58.3 & 59.1 \\ 
  110 & 110 & 169 &  82.0 \textcolor{green}{\scriptsize (+2.6)} & 68.3 \textcolor{red}{\scriptsize (-1.2)} & 59.1 \textcolor{green}{\scriptsize (+0.8)} & 60.4 \textcolor{green}{\scriptsize (+1.3)} \\
  440 & 440 & 64 & 81.6 \textcolor{green}{\scriptsize (+2.2)} & 67.2 \textcolor{red}{\scriptsize (-2.3)} & 59.4 \textcolor{green}{\scriptsize (+1.1)}  & 60.2 \textcolor{green}{\scriptsize (+1.1)} \\
    \bottomrule
    \end{tabular}%
\label{tab:frame_ablation}
\end{table*}

\begin{table*}[t]
\tabstyle{1pt}
\centering
\small
\caption{Comparison of different video representations. The video representation $\Vcal$ is consistent in training and inference for all methods, except that SlowFast-LLaVA considers simple representation $\Vcal$ in training and its specified $\Vcal$ in inference.}
\begin{tabular}{@{}lll|llll@{}}
    \toprule
       & & & \multicolumn{2}{c}{in-domain} & \multicolumn{2}{c}{out-of-domain} \\   
        \multirow{3}{*}{Method}  &
      \multirow{3}{*}{$\Vcal=(T,M,s,p)$}  & \#Visual &
    \textbf{NExT-QA} & \textbf{PerceptionTest} & 
    \textbf{EgoSchema} & 
    \textbf{VideoMME} \\ \cmidrule(l){4-7} 
      &  & Tokens& mc & val  & test & wo \\ \midrule
    Simple representation & (32, 729, 1, 2) & 5,408 & 80.4 & 68.2 & 56.3 & 59.1 \\
     \MethodName{} & (64, 729, 3, 2) & 5,396 & 81.1   & 67.7  & 57.1  & 59.8  \\ 
     LITA &
    (42, 729, 2, 2)& 5,313  & 80.8 & 68.3 & 54.3 & 59.1 \\
     SlowFast-LLaVA & (42, 729, 2, 2) & 5,313 & 79.4   & 68.2  & 56.2  & 58.9 \\     
    \bottomrule
    \end{tabular}%
\label{tab:video-bench-slow-fast}
\end{table*}

\section{Beyond Singularity: Extensive Sampling Matters}

We perform experiments to explore how video representations affect the model's performance. All experiments were carried out in a video-only setting, using video data with durations from 0 to 30 seconds as our training data. We focused on evaluating how the number of frames and the number of visual tokens per frame impact model performance. Regarding the frame count, it is noteworthy that observing the effects of a high number of frames---such as over 100---does not necessarily require long videos. Our results indicate that the dynamic properties of the data render even 100 frames insufficient to fully capture the condent of a 30-second video, which typically runs at 15 FPS.

In Table~\ref{tab:frame_ablation}, the first group shows an increase in the number of frames from 32 to 110. We set 110 frames as the upper limit to avoid overloading the GPU. With more frames, we see significant improvements in all datasets. While it's generally expected that using more frames boosts performance, previous studies~\citep{luo2021clip4clip,lei2021less,lei2022revealing} have noted that performance tends to plateau when training with more than 16 frames. We propose that the saturation observed in earlier studies arises due to the selection of training datasets such as MSVD~\citep{chen2011msvd} and WebVid~\citep{Bain21}, where the video content is highly static, allowing a small number of frames to represent the entire video effectively. In contrast, the dynamic nature of the videos and the detailed nature of the annotations in~\DataName{} allow for continuous benefits from extensive sampling

The second group in Table~\ref{tab:frame_ablation} demonstrates the effects of varying the number of inference frames while keeping the number of training frames constant. A modest increase in the inference frames slightly enhances performance; however, excessively increasing the number of inference frames can degrade it.

In Table~\ref{tab:frame_ablation}'s third group, we illustrates the trade-off between the number of frames and the number of tokens per frame. Configurations with fewer tokens per frame but more frames yield superior results, even with a lower total count of visual tokens (18,590 versus 21,632). This finding emphasizes that increasing the number of frames, rather than the tokens per frame or the total number of tokens, enhances performance. However, a balance is necessary; as the number of frames increases to 440 and the tokens per frame decreases to 64, performance drops. This observation led us to use \MethodName{} for video representation.

\section{Capabilities}
Beyong achieve good benchmark performance, Our observations of \ModelName{} reveal various capabilities in video
understanding. Specifically, it show a great abilities in the understanding video using real-world knowledge,including, but not limited to:

\begin{itemize}[leftmargin=7.5mm]
\setlength{\itemsep}{2pt}
\item {\it Optical Illusion}: As shown in Table~\ref{tab:drawing}, \ModelName{} recognizes that the green dragon in the video is not a real 3D object. It appears three-dimensional due to an optical illusion that affects human perception.
\item {\it Special Domain}: As indicated in Table~\ref{tab:drawing}, \ModelName{} understands the content within special domains in the video, such as sketches and fights in video games.
\item {\it Unusual Action}: As detailed in Table~\ref{tab:unusual_action}, \ModelName{} identifies atypical actions in the video, such as "physical therapy" for pets, beyond ordinary activities.
\item {\it Physical Laws}: As shown in Table~\ref{tab:physical_law}, \ModelName{} comprehends basic physical laws demonstrated in the video, like zero gravity in space stations, which allows objects to float without falling.

\end{itemize}

\begin{table*}[htp]
\begin{minipage}{0.99\textwidth}
\begin{AIbox}{Real-world Knowledge: Illusion}
\centering
\scalebox{0.75}{
\begin{tabular}{l p{13.5cm}}
&  \includegraphics[width=13.5cm]{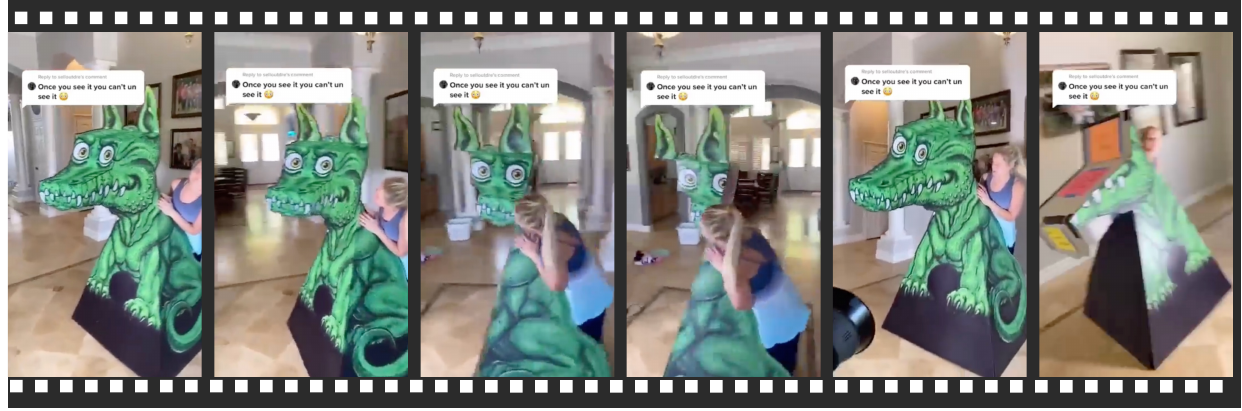} \\
\footnotesize
User & Please describe this video. \\
\midrule
\footnotesize
\ModelName{} & The video features a person interacting with a large, cartoonish green dragon cutout in a well-lit, spacious room adorned with framed pictures on the walls. The person, dressed in a blue tank top and light blue jeans, is seen hugging and adjusting the dragon cutout, which has a wide, toothy grin and large, expressive eyes. The text overlay reads 'Reply to seloutria's comment: Once you see it you can't un see it,' suggesting that the cutout is designed to be visually striking and memorable. The person moves around the cutout, showcasing different angles and perspectives. \\ \midrule
\footnotesize
User & Is this green dragon a real 3D structure? \\
\midrule
\footnotesize
\ModelName{} & No, the green dragon is a 2D cutout that creates an optical illusion. \\
\end{tabular}
}
\end{AIbox}
\captionof{table}{\ModelName{} learns to understand the \textit{optical illusion} in the video.}
\label{tab:dragon}
\end{minipage}
\end{table*}

\begin{table*}[htp]
\begin{minipage}{0.99\textwidth}
\begin{AIbox}{Real-world Knowledge: Special Domain}
\centering
\scalebox{0.75}{
\begin{tabular}{l p{13.5cm}}
&  \includegraphics[width=13.5cm]{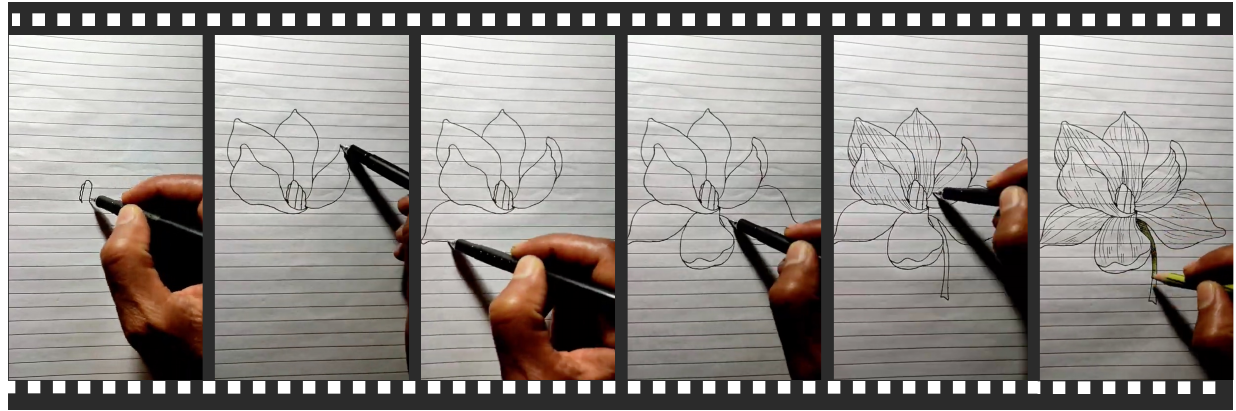} \\
\footnotesize
User &Please describe this video. \\
\midrule
\footnotesize
\ModelName{} & The video begins with a blank, lined sheet of paper. A hand holding a black pen starts drawing on the paper, creating the outline of a flower with five petals. The artist adds details to the petals, including lines and shading to give them a more realistic appearance. The stem and leaves are also drawn, completing the flower illustration. The artist adds final touches to the flower drawing, enhancing the details and shading to give it a more realistic appearance. The artist then switches to a yellow pen and begins coloring the center of the flower, adding depth and contrast to the drawing. The video concludes with the artist adding final touches to the coloring, enhancing the overall appearance of the flower illustration. \\
\midrule
\footnotesize
User & What could this flower be? \\
\midrule
\footnotesize
\ModelName{} & This flower could be a magnolia or a similar species. \\
\bottomrule
&  \includegraphics[width=13.5cm]{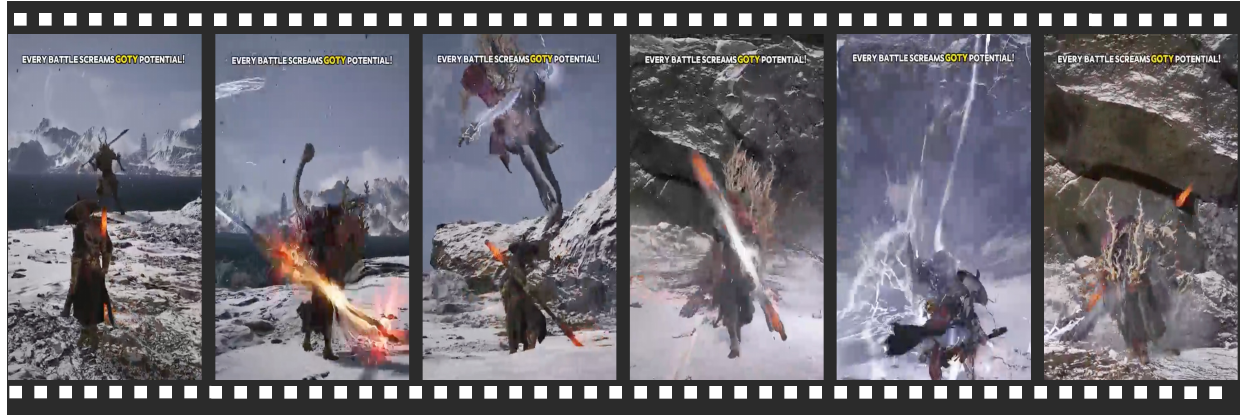} \\
\footnotesize
User &Please describe this video. \\
\midrule
\footnotesize
\ModelName{} & The video showcases an intense battle scene set in a snowy, mountainous landscape. A character dressed in dark, heavy armor with glowing orange accents wields a sword and faces off against a large, menacing creature with glowing red eyes and antlers. The text 'EVERY BATTLE SCREAMS GOTY POTENTIAL!' is prominently displayed at the top of each frame, emphasizing the high stakes and epic nature of the confrontation.

The battle is fierce, with the character and the creature exchanging powerful attacks. The creature emits blue energy and lightning, while the character retaliates with fiery and icy attacks. The snowy ground is littered with debris and frost, adding to the harsh and brutal environment. The background features a large body of water and a distant lighthouse, enhancing the sense of isolation and danger. \\
\end{tabular}
}
\end{AIbox}
\captionof{table}{\ModelName{} learns to understand the the video in \textit{special domain}.}
\label{tab:drawing}
\end{minipage}
\end{table*}


\begin{table}[htp]
\begin{minipage}{0.99\textwidth}
\begin{AIbox}{Real-world Knowledge: Unusual Action}
\centering
\scalebox{0.75}{
\begin{tabular}{l p{13.5cm}}
&  \includegraphics[width=13.5cm]{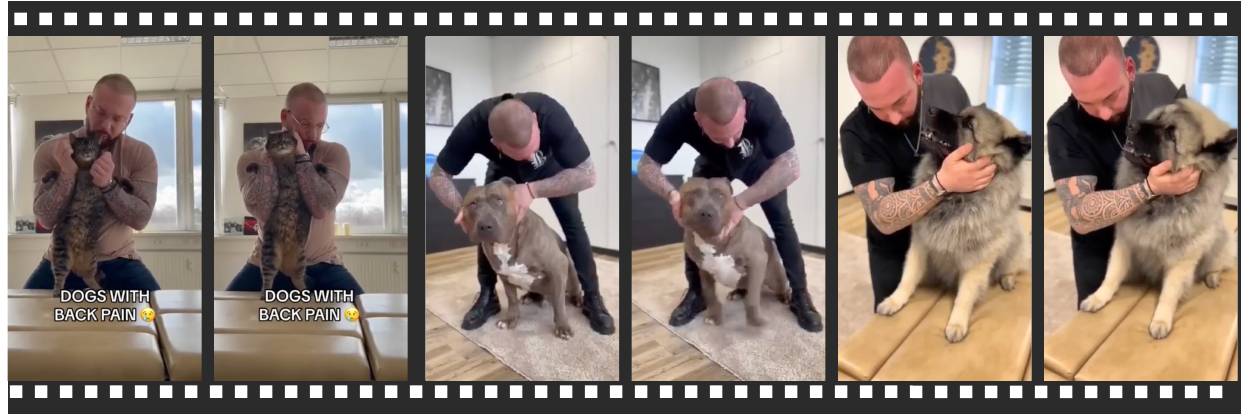} \\
\footnotesize
User & Please describe this video.\\
\midrule
\footnotesize
\ModelName{} & The video begins with a person holding a tabby cat, with the text 'DOGS WITH BACK PAIN' and a crying emoji displayed at the bottom of the screen. The person, who has tattoos on their arms and is wearing a light-colored shirt, is in a room with large windows and a white wall. The scene transitions to the person petting a large, brown and white dog, who is sitting on the floor. The person then stands up and gestures with their hands, while the dog looks up at them. The next scene shows the person kneeling and petting a large, fluffy dog with a black and white coat, who is sitting on a table. The person continues to pet the dog, who appears calm and relaxed. \\ \midrule
\footnotesize
User & What did this man do to his pets?\\
\midrule
\footnotesize
\ModelName{} & This man provided physical therapy to his pets. \\
\end{tabular}
}
\end{AIbox}
\captionof{table}{\ModelName{} learns to understand the \textit{unusual action} in the video.}
\label{tab:unusual_action}
\end{minipage}
\end{table}


\begin{table*}[htp]
\begin{minipage}{0.99\textwidth}
\begin{AIbox}{Real-world Knowledge: Physical Laws}
\centering
\scalebox{0.75}{
\begin{tabular}{l p{13.5cm}}
&  \includegraphics[width=13.5cm]{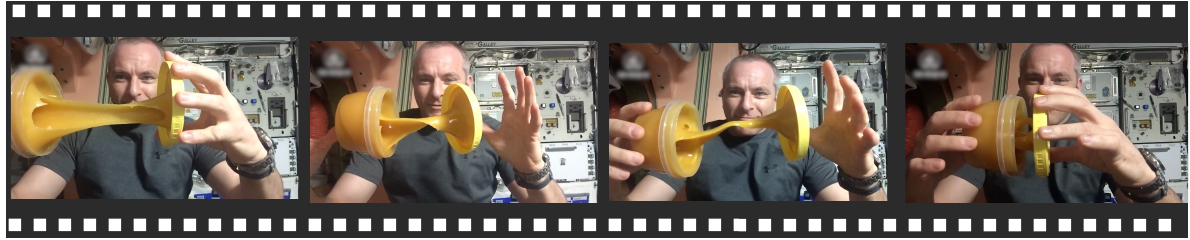} \\
\footnotesize
User &Please describe this video. \\
\midrule
\footnotesize
\ModelName{} & The video features a person in a space station environment, wearing a dark gray shirt and a wristwatch. They are holding two containers filled with a thick, yellowish substance. The person tilts the containers to demonstrate the fluid's behavior, showing how it stretches and moves between the containers. The background includes various equipment and panels typical of a space station, with labels such as 'GALLEY' and 'FUSION'. The person continues to manipulate the containers, emphasizing the fluid's properties. The video concludes with the person holding one container and twisting the lid to close it, further demonstrating the fluid's properties. \\
\midrule
\footnotesize
User & Why doesn't this yellowish substance fall? \\
\midrule
\footnotesize
\ModelName{} & This yellowish substance doesn't fall because it is in zero gravity.\\\end{tabular}
}
\end{AIbox}
\captionof{table}{\ModelName{} learns to understand the \textit{physical laws} in the video.}
\label{tab:physical_law}
\end{minipage}
\end{table*}

\end{document}